%% file: main.tex
\pdfoutput=1
\documentclass[11pt]{article}

\usepackage[final]{acl}

\usepackage{times}
\usepackage{latexsym}

\usepackage[T1]{fontenc}
\usepackage[utf8]{inputenc}

\usepackage{microtype}

\usepackage{inconsolata}

\usepackage{graphicx}

\usepackage{algorithm}
\usepackage{algorithmic}
\usepackage{amssymb}
\usepackage{amsmath,amsfonts}
\usepackage{booktabs}
\usepackage{multirow}
\usepackage{pifont}
\usepackage{makecell}

\usepackage{colortbl}
\PassOptionsToPackage{table,xcdraw}{xcolor}
\usepackage[normalem]{ulem}
\useunder{\uline}{\ul}{}
\usepackage{mdframed}
\usepackage{xurl}

\DeclareFixedFont{\ttb}{T1}{txtt}{bx}{n}{8} % for bold
\DeclareFixedFont{\ttm}{T1}{txtt}{m}{n}{8}  % for normal
\usepackage{color}
\definecolor{deepblue}{rgb}{0,0,0.5}
\definecolor{deepred}{rgb}{0.6,0,0}
\definecolor{deepgreen}{rgb}{0,0.5,0}
\definecolor{codegray}{rgb}{0.5,0.5,0.5}
\usepackage{listings}
\newcommand\pythonstyle{\lstset{
language=Python,
basicstyle=\ttm,
commentstyle=\color{codegray}\fontsize{8pt}{10pt}\selectfont,
morekeywords={self},              % Add keywords here
keywordstyle=\ttb\color{deepblue},
emph={MyClass,__init__},          % Custom highlighting
emphstyle=\ttb\color{deepred},    % Custom highlighting style
stringstyle=\color{deepgreen},
frame=tb,                         % Any extra options here
breaklines=true,                  % Enable line breaking
breakatwhitespace=true,
showstringspaces=false
}}
\lstnewenvironment{python}[1][]
{
\pythonstyle
\lstset{#1}
}
{}

\newcommand\pythoninline[1]{{\pythonstyle\lstinline!#1!}}

\title{AMXFP4: Taming Activation Outliers with\\ Asymmetric Microscaling Floating-Point for 4-bit LLM Inference}

\author{Janghwan Lee${}^{1}$, Jiwoong Park${}^{1}$, Jinseok Kim${}^{2}$, Yongjik Kim${}^{2}$, \\
        {\bf Jungju Oh${}^{2}$}, {\bf Jinwook Oh${}^{2}$} and {\bf Jungwook Choi${}^{1\dagger}$} \\
        \normalsize{\textsuperscript{1}Hanyang University},
        \normalsize{\textsuperscript{2}Rebellions Inc.} \\
        \small{\textsuperscript{1}\texttt{\{hwanii0288, pjw9703\}@hanyang.ac.kr}} \\
        \small{\textsuperscript{2}\texttt{\{jinseok, yongjik.kim, jungju.oh, j.oh\}@rebellions.ai}, \textsuperscript{1$\dagger$}\texttt{choij@hanyang.ac.kr}}%
}

\begin{document}
\maketitle

\renewcommand{\thefootnote}{}  % disable numbering
\footnotetext{${}^\dagger$\,Corresponding author.}
\renewcommand{\thefootnote}{\arabic{footnote}}

\begin{abstract}

As large language models (LLMs) grow in parameter size and context length, computation precision has been reduced from 16-bit to 4-bit to improve inference efficiency. However, this reduction causes accuracy degradation due to activation outliers. Rotation-based INT4 methods address this via matrix calibration, but they introduce multi-hour overheads and leave key computations in full precision. Microscaling (MX) floating-point (FP) formats offer fine-grained representation with a shared scale, enabling fully quantized matrix multiplications through direct casting without calibration. However, existing research shows unsatisfactory empirical results for MXFP4 inference, and the robustness of MX formats remains largely unexplored. 

In this work, we uncover the fundamental tradeoffs of the MX format: while it effectively suppresses activation outliers, it does so at the cost of increased group-wise asymmetry. To address this, we propose AMXFP4, a 4-bit asymmetric FP format that handles both issues using asymmetric shared scales, without requiring calibration. Our custom MAC engine adds negligible hardware cost while improving accuracy: AMXFP4 outperforms MXFP4 by 3\% on VQA and exceeds rotation-based methods by 1.6\% on CSQA. It also surpasses recently deployed commercial MXFP4 variants. Code: \url{https://github.com/aiha-lab/MX-QLLM}

\end{abstract}

\input{Sections/introduction}

\input{Sections/background}

\input{Sections/llm_outlier}

\input{Sections/rp_format}

\input{Sections/experiments}

\input{Sections/conclusion}

\section*{Limitations}

While AMXFP4 shows strong promise across various LLM tasks, our current hardware analysis remains focused on a MAC-level evaluation. This choice reflects a balanced starting point for proof-of-concept experiments and aligns with many common practices in precision-scaling research~\cite{darvish2023shared}. However, as seen with recent system-level benchmarks (e.g., NVIDIA’s Blackwell), there is significant potential to extend these findings to a full system-level evaluation. We plan to extend our evaluation accordingly, examining factors such as overall throughput, energy efficiency, and system-level trade-offs.

Additionally, our experiments have employed greedy decoding to ensure fair comparisons. However, recent deployment scenarios often rely on more advanced strategies—such as best-of-N sampling or self-refinement in reasoning LLMs—which require increased computational resources at inference time. Investigating AMXFP4’s robustness and efficiency under these test-time scaling conditions is a natural next step and could further underscore the method’s potential benefits in real-world applications.

\bibliography{custom}

\appendix
\input{Sections/appendix-mx}

\input{Sections/appendix-experiments}

\end{document}

%% file: Sections/introduction.tex
\section{Introduction}
\label{sec1}

\begin{figure*}[t]
\centering
\centerline{\includegraphics[width=\textwidth]{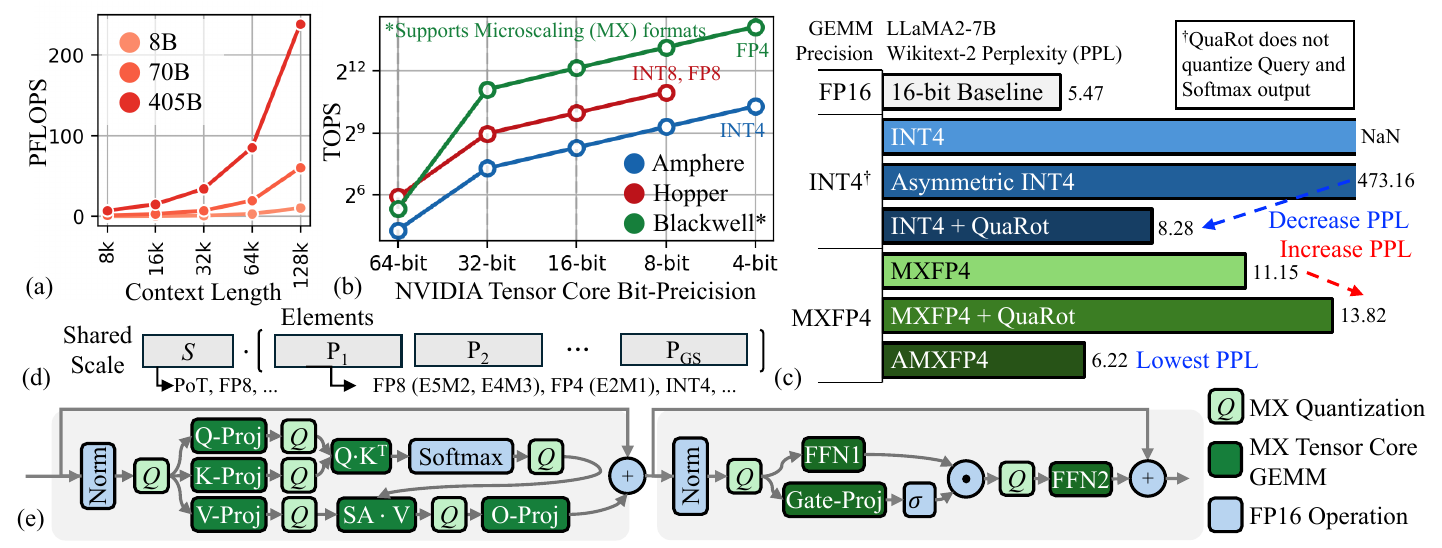}}
\vspace{-2mm}
\caption{(a) FLOPS across context length and model sizes. (b) Precision scaling in NVIDIA Tensor Cores. (d) Impact of bit-precision and data rotation on perplexity. (d) MX format. (e) LLM inference with MX Tensor Core.}
\label{fig:mx}
\end{figure*}

Multi-modal Large Language Models (LLMs) are widely used in advanced natural language processing tasks, including chatbots, long-document question-answering, and visual graph interpretation~\cite{bai2023qwen, liu2023llava}. To enhance their capabilities, LLMs have been significantly scaled in both parameter size and context length~\cite{chung2022scaling,chowdhery2022palm}. For example, LLaMA3~\cite{llama3modelcard} now features 405 billion parameters and supports context lengths of up to 128K tokens. As shown in Fig.~\ref{fig:mx}(a), this scaling results in peta-FLOP-level computational demands during the prefill phase, where the model processes user context before inference.

Leading computing platforms have focused on bit-precision scaling to meet the computational demands of LLMs~\cite{h100,b100,maia}. Reducing operand bit-widths improves area and energy efficiency in arithmetic operations~\cite{horowitz2014energy}, enabling higher computation density in accelerators. As shown in Fig.~\ref{fig:mx}(b), NVIDIA’s Tensor Cores double computation speed by lowering multiply-accumulate (MAC) precision from FP16 to FP8~\cite{h100} and from INT8 to INT4~\cite{a100}.

Recent research explores activation and weight quantization to improve LLM inference efficiency by leveraging hardware precision scaling. However, quantizing both weights and activations to INT4 often degrades accuracy due to activation outliers~\cite{dettmers2022llmint8,xiao2022smoothquant}. Rotation-based transformations mitigate this by making activations more quantization-friendly~\cite{ashkboos2024quarot,liu2024spinquant}, with approaches like QuaRot~\cite{ashkboos2024quarot} significantly reducing LLM perplexity in INT4 inference (Fig.~\ref{fig:mx}(c)). Despite these benefits, rotation-based methods require extensive calibration, leading to overfitting risks~\cite{enhancing,lin2023awq} (cf. Table~\ref{tab:overfitting}), and are impractical for user-specific model deployments that demand frequent recalibration~\cite{bang-etal-2024-crayon}. Additionally, they leave Softmax outputs unquantized, forcing FP16 multiplications with value vectors, which account for 41\% of total FLOPs in 8B LLMs with 128K-token inputs (cf. Fig.~\ref{fig:rotation}).

An alternative approach to quantization introduces reduced-precision formats that enable calibration-free data-type conversion (i.e., direct casting). For instance, the latest NVIDIA Tensor Core~\cite{b100} supports the microscaling (MX) format, introduced by the Open Compute Project (OCP)~\cite{ocp}, which groups low-precision elements under a shared scale to mitigate dynamic range limitations (Fig.~\ref{fig:mx}(b), (d)). As shown in Fig.~\ref{fig:mx}(c), (e), MXFP4 achieves full matrix quantization with minimal perplexity degradation compared to INT4, without requiring data rotation. This is due to its fine-grained quantization, which enhances value representation precision. However, MXFP4 still lags behind the 16-bit baseline in perplexity and performs worse when combined with data rotation, and the root causes of this destructive interaction are mainly unexplored.

This work uncovers a key trade-off in the MX format: while it effectively suppresses activation outliers, it increases group-wise asymmetry. Grouping activation tensors into small micro-scaled units mitigates outliers, similar to rotation methods, but enables direct-cast inference. However, this grouping amplifies data asymmetry, necessitating an asymmetric numerical representation. To address this, we propose AMXFP4, a microscaling floating-point format designed for robust 4-bit LLM inference, which effectively handles activation outliers through micro-scaled asymmetric data representation. By employing an FP8 shared scale for both weights and activations, AMXFP4 achieves quantization error rates close to ideal Lloyd-Max quantization. To validate its broad applicability, we evaluate AMXFP4 across multi-turn conversation, long-context inference, and visual question-answering (VQA) tasks on decoder-only LLMs, vision-language models, and an encoder-decoder model. Results show that AMXFP4 enables calibration-free, direct-cast 4-bit inference, outperforming MXFP4 and leading rotation-based quantization methods. Additionally, AMXFP4 performs better than the recently deployed commercial MXFP4 format (NVFP4)~\cite{tensorrt-llm}.

Our contributions can be summarized as follows:
\begin{itemize}
    \item We examine the MXFP4 format, finding that microscaling effectively reduces activation outliers without calibration but introduces asymmetry, necessitating asymmetric numerical representation.
    \item We propose AMXFP4, a novel format that combines FP4 elements with shared asymmetric FP8 scales, significantly suppressing quantization error.
    \item We evaluate AMXFP4 across diverse applications, including multi-turn conversation, long-context inference, and VQA, across multiple model types, demonstrating consistently superior accuracy to MXFP4.
\end{itemize}

%% file: Sections/background.tex
\section{Background and Related Work}

\subsection{Bit-Precision Scaling for Accelerators} 
\label{sec2.1}
Reduced-precision formats are vital for enhancing scalability and computational efficiency in deep learning accelerators, conserving area and energy in direct proportion to bit-width reduction~\cite{horowitz2014energy}. This scaling enables higher floating-point operations per second (FLOPS) with lower power usage, thereby increasing accelerator throughput. For instance, NVIDIA’s Tensor Cores have progressed from FP16 in Volta~\cite{v100} to FP8 in Hopper~\cite{h100} and FP4 in Blackwell~\cite{b100}, boosting computational speeds from 112 tera to 20 peta FLOPS, as shown in Fig.~\ref{fig:mx}(b). Similar advancements by other computing platform companies in scaling precision from 16-bit to 4-bit are crucial for managing the growing complexity of LLMs~\cite{amd_mi325,maia}. 

Recently, the microscaling (MX) format~\cite{ocp, darvish2023shared, rouhani2023microscaling} has been developed from Block Floating Point (BFP)~\cite{drumond2018training,darvish2020pushing} by incorporating a shared scale across a block of reduced-precision elements, thus mitigating quantization error due to limited dynamic range. While the original BFP format allows flexibility in design parameters-exponent ($E$) and mantissa ($M$) for the element ($P_i$) and the shared scale ($S$), and the group size ($GS$), MX prescribes specific \textit{MX-compliant} configurations (cf. Table~\ref{tab:mx_configs}): MXFP8 ($P_i$:$E4M3$, $S$=$E8$, $GS$:$32$) and MXFP4 ($P_i$:$E2M1$, $S$=$E8$, $GS$:$32$), as shown in Fig.~\ref{fig:mx}(d).

However, MXFP4's robustness for LLM inference remains uncertain, with significant performance degradation in 4-bit inference due to activation quantization~\cite{rouhani2023microscaling}. Moreover, MXFP4 lacks validation on practical tasks such as multi-turn chatbot interactions, raising concerns about its real-world applicability. While MXFP4 models generate coherent answers, they often yield unhelpful responses, consistent with findings that quantization can impair conversational quality~\cite{lee-etal-2024-improving-conversational} (e.g., Fig.~\ref{fig:chat-example}). These results underscore the need for new data formats to enable robust 4-bit inference.

\subsection{Quantizing LLM's Activation and Weight}
\label{sec2.2}

Recent research highlights the difficulty quantifying LLM activations due to outliers extending the activation dynamic range, leading to increased quantization error~\cite{xiao2022smoothquant,ashkboos2024quarot}. Prior studies propose rescaling weights and activations to reshape their distributions for better quantization compatibility while preserving mathematical equivalence~\cite{xiao2022smoothquant,shao2024omniquant,enhancing}. However, such methods often experience accuracy degradation in 4-bit inference~\cite{lin2024duquantdistributingoutliersdual}. Data rotation strategies, including QuaRot~\cite{ashkboos2024quarot} and SpinQuant~\cite{liu2024spinquant}, use orthogonal matrices to redistribute concentrated channel information (represented as $R$ in Fig.~\ref{fig:rotation}(a)). QuaRot applies a randomized Hadamard matrix, while SpinQuant uses learned rotation matrices. DuQuant further enhances this approach by combining per-channel permutation and rotation, achieving state-of-the-art accuracy in 4-bit inference~\cite{lin2024duquantdistributingoutliersdual}.

However, these rotation-based methods exclude quantization for the Softmax output, leaving matrix multiplications in the self-attention calculation to be computed in FP16. Since self-attention computation scales quadratically with context length during the prefill phase, the partial quantization of rotation methods significantly reduces overall computational efficiency in long-context inference. Additionally, these techniques require extensive calibration, such as GPTQ~\cite{frantar2022gptq} or training rotation matrices, to improve model accuracy. However, calibration introduces the risk of overfitting, as models may become overly tailored to the calibration dataset, limiting their adaptability across broader applications (Table~\ref{tab:overfitting}). Further discussions on limitations of calibration-based methods are provided in the Appendix~\ref{appendix:rotation}.

These challenges highlight the need for a generalizable quantization approach that minimizes calibration dependence and applies uniformly across computations. Although MXFP4, a previously explored reduced-precision format, applies to all matrix multiplication without calibration, it compromises model accuracy. This work analyzes MXFP4’s strengths and limitations, and proposes AMXFP4, a superior 4-bit format that enables direct-casting with improved model accuracy.

%% file: Sections/llm_outlier.tex
\section{Microscaling for Taming Outliers}
\label{sec3}

We systematically analyze activation outliers across various LLMs using representative statistical measures—kurtosis and mean—to understand the effects of microscaling (i.e., reducing a quantization group to 32 elements). Kurtosis, the fourth standardized moment, is commonly used to assess the prevalence of outliers~\cite{liu2024spinquant}, while the mean reflects asymmetry within each group. We use box plots of kurtosis and mean to examine the value distribution within groups, which are subject to quantization using a shared scale.

\begin{figure}[t]
\centering
\centerline{\includegraphics[width=\columnwidth]{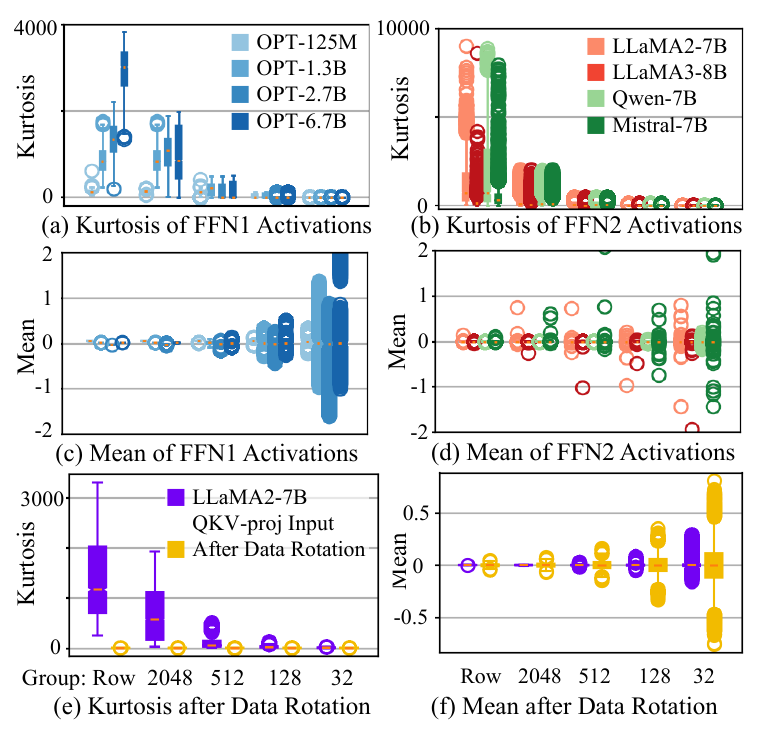}}
\vspace{-3mm}
\caption{Data characteristics based on (a-d) types of LLM, layer and (e-f) data rotation across group sizes.}
\label{fig:model_stats}
\end{figure}

\subsection{Analysis of LLM's Activation Outliers} 
\label{sec3.1}

Fig.~\ref{fig:model_stats}(a) and (b) present the kurtosis box plots for the OPT~\cite{opt} and LLaMA-like models (LLaMA, Qwen, Mistral~\cite{touvron2023llama2,llama3modelcard,bai2023qwen,jiang2023mistral}). In cases of row-wise grouping (typically $GS\gg1024$), the OPT models exhibit high kurtosis in FFN1 activations, indicating many outliers that challenge quantization. Additionally, outlier prevalence increases with model size, aligning with previous findings that larger models are more affected by quantization~\cite{dettmers2022llmint8}. Conversely, LLaMA-like models use the Gated Linear Unit (GLU) activation function, involving extra matrix multiplication; thus, data passing through FFN1 undergoes element-wise multiplication before FFN2, further amplifying outliers—a phenomenon observed in recent studies~\cite{yang2024mitigatingquantizationerrorsactivation,fishman2024scalingfp8trainingtrilliontoken}. Notably, \textit{outlier dominance is reduced as group size decreases in both model types}. At $GS$=$32$, kurtosis nearly disappears, suggesting the activation dynamic range within groups becomes more suitable for quantization. This observation helps explain the preliminary success of MXFP8 in direct-casting for selected LLMs~\cite{rouhani2023microscaling}, but it does not explain the disappointing performance of MXFP4. 

To assess the trade-offs in the MX format's handling of outliers, we examine the box plots of group means, which reflect distribution asymmetry. Fig.~\ref{fig:model_stats}(c) and (d) show the mean values for FFN1 and FFN2 input activations as group size decreases from an entire row to 32. Notably, with large group sizes, group means center around zero, but as group size decreases, the means scatter significantly. This scattering indicates that the symmetric data representation typically used in the MX format is suboptimal for microscaled activation quantization. In other words, \textit{microscaling addresses activation outliers at the cost of data symmetry}. Thus, simply reducing group size (as in the MX format) may not adequately minimize quantization error; instead, an asymmetric data representation becomes essential.

\subsection{Data Rotation vs. Microscaling} 
\label{sec3.2}

We then examine how data rotation reduces outliers alongside microscaling and assess its effectiveness as group size decreases. Fig.~\ref{fig:model_stats}(e) shows the kurtosis before and after applying data rotation using a random Hadamard transform~\cite{ashkboos2024quarot} across decreasing group sizes. When the group size spans an entire row, activation rotation substantially lowers kurtosis, demonstrating its efficacy in 4-bit activation quantization. However, as group size decreases, the original activation’s kurtosis also drops, reaching levels comparable to those achieved with rotation. Thus, the benefit of data rotation in outlier reduction diminishes with smaller group sizes.

On the other hand, Fig.~\ref{fig:model_stats}(f) shows the group means of the activation before and after applying data rotation. As with the original activation, the group means scatter more as group sizes decrease, but this scattering is even more pronounced with rotated activations. This indicates that rotation introduces an additional asymmetry in group distributions, which complicates quantization with MXFP4's symmetric representation (cf. Table~\ref{tab:transformation}). In other words, data rotation and microscaling lack synergy, as both focus on outlier suppression without addressing asymmetry. Thus, a microscaling data format that effectively handles group distribution asymmetry presents a compelling alternative.

\begin{figure}[t]
\centering
\centerline{\includegraphics[width=\columnwidth]{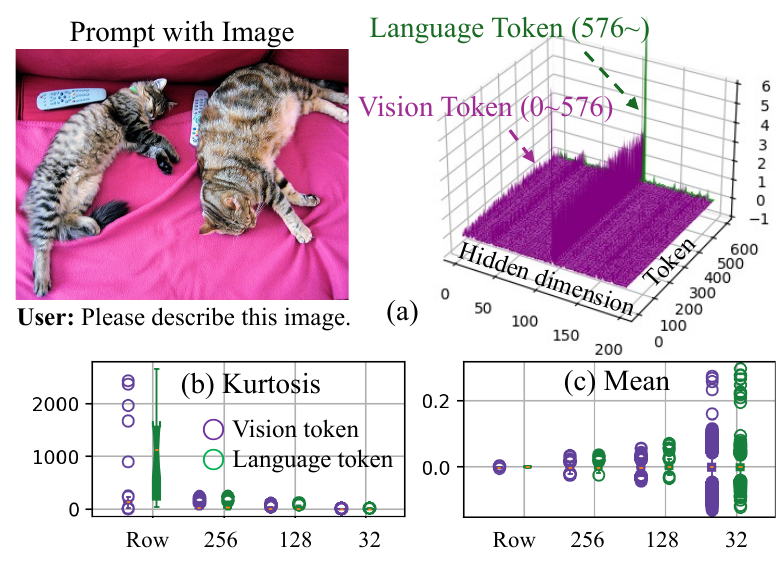}}
\vspace{-2mm}
\caption{Characteristic of VLM activation outliers across group sizes (LLaVA-v1.6-Vicuna-7B Layer 1 QKV-Proj).}
\label{fig:vlm-outliers}
\end{figure}

\subsection{Multi-modal LLM’s Activation Outlier}
\label{sec3.3}

To further understand activation outliers under microscaling in multi-modal LLMs, we examine the popular vision-language model LLaVA~\cite{liu2023llava}. LLaVA combines a visual encoder and a language model backbone: an image is processed by a vision transformer-based encoder~\cite{dosovitskiy2021an_vit} to generate vision tokens, which are then input to the language model along with language tokens from the user prompt.

As shown in Fig.~\ref{fig:vlm-outliers}(a), both vision and language tokens exhibit outliers within the same hidden dimension of the activation, though their distributions differ. Language tokens typically concentrate around larger magnitudes, while only some vision tokens reach high magnitudes, a trend observed consistently across layers. In Fig.~\ref{fig:vlm-outliers}(c), these differences result in varying kurtosis distributions for row-wise group quantization: language tokens have clustered outliers, while vision tokens show a sparser outlier distribution. However, this distinction fades as group size decreases, illustrating the effectiveness of microscaling in suppressing outliers. Similar to LLMs, LLaVA's group means scatter as group size decreases, indicating increased asymmetry in exchange for outlier suppression. This suggests microscaling could better handle diverse outlier patterns from vision and language tokens if designed to support asymmetric data representation.

%% file: Sections/rp_format.tex
\section{Asymmetric Microscaling Format}
\label{sec4}
The findings from Sec.~\ref{sec3} motivate the development of a new microscaling format that inherently supports asymmetric data representation. In this section, we explore the design space of the microscaling data format ($P_i$ and $S$) alongside considerations for asymmetric quantization schemes.

\subsection{Selecting Element-Wise Data Format}
\label{sec4.1}
We first examine the design space of the element-wise data format $P_i$. To evaluate the benefits of asymmetric formats, we compare the mean-square error (MSE) on activation samples from LLaMA2-7B’s QKV-Proj at layer 5 across four symmetric formats (INT4, FP4, NF4~\cite{dettmers2023qlora}, SF4~\cite{dotzel2024students}) with two asymmetric formats:

\begin{itemize}
    \item \textbf{Asymmetric INT (AsymINT):} INT quantization applies asymmetry through a zero-point, shifting the data range from zero-centered to span between the minimum and maximum values~\cite{dettmers2022llmint8}.
    \item \textbf{Asymmetric FP (AsymFP):} FP quantization introduces asymmetry by applying separate scales to positive and negative values due to FP’s inherently zero-centered representation~\cite{zhang2023afpq}.
\end{itemize}

\begin{figure}[t]
\centerline{\includegraphics[width=\columnwidth]{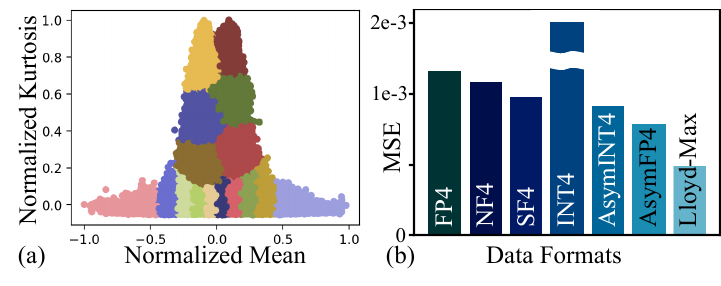}}
\vspace{-3mm}
\caption{Cluster-wise Lloyd-Max quantization and quantization error across data formats (LLaMA2-7B layer 5 QKV-Proj input activation). Detailed cluster-wise error statistics and results from other layers are provided in Table~\ref{tab:cluster-mse} and \ref{tab:layerwise-mse}.}
\label{fig:lloyd-max}
\end{figure}

We compare the MSE of each format on activation samples from LLaMA2-7B’s QKV-Proj at layer 5. Fig.~\ref{fig:lloyd-max}(a) characterizes these activations by group mean (x-axis) and kurtosis (y-axis). As a reference, we cluster groups based on mean and kurtosis similarity, then apply the Lloyd-Max algorithm~\cite{lloydmax} for near-optimal quantization (100 iterations, with 16 clusters, as further clustering yields no additional MSE reduction).

Fig.~\ref{fig:lloyd-max}(b) presents the MSE of various element-wise data formats. Compared to Lloyd-Max quantization (used as a reference), all symmetric data formats show a significant MSE increase, with INT4 experiencing the most notable degradation. In contrast, AsymINT4 and AsymFP4 achieve lower MSE, with AsymFP showing MSE closest to Lloyd-Max (a consistent trend across models and layers). This finding supports the selection of AsymFP4 as the element-wise format, further validated empirically in Table~\ref{tab:transformation}.

\begin{figure}[t]
\centerline{\includegraphics[width=\columnwidth]{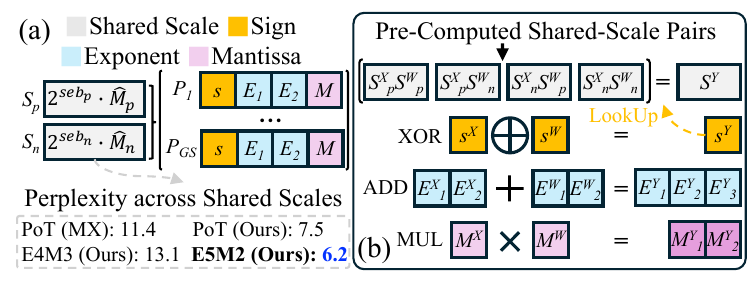}}
\caption{(a) Illustration of AMXFP4 and LLaMA2-7B Wikitext-2 perplexity across shared scale types. (b) Multiplication between two AMXFP4 datas.}
\label{fig:amxfp4}
\end{figure}

\subsection{Selecting Shared-Scale with Asymmetry}
\label{sec4.2}
With AsymFP4 selected as the preferred element-wise data representation, its original design for weight-only quantization~\cite{zhang2023afpq} requires high-precision dequantization before multiplication with activations. To integrate AsymFP into reduced-precision GEMM, we redefine AsymFP such that an exponent-bit-shifted mantissa represents a value, which is then scaled by a shared factor with sign-dependent polarity:
\begin{align}
\label{eq:asymfp}
  x_q &= 
  \begin{cases}
     (-1)^{s}\cdot 2^{E+eb}\cdot{M}\cdot (2^{seb_p}\cdot{\hat{M_p}}) \text{ if } s=0, \\
     (-1)^{s}\cdot 2^{E+eb}\cdot{M}\cdot (2^{seb_n}\cdot{\hat{M_n}}) \text{ if } s=1, \\
  \end{cases} 
\end{align}
where $s$, $E$, $eb$, and $M$ represent an element's sign, exponent, exponent bias, and mantissa, respectively. As described in Fig.~\ref{fig:amxfp4}(a), the terms $2^{seb_p}\cdot{\hat{M_p}}$ and $2^{seb_n}\cdot{\hat{M_n}}$ represents the positive and negative scales shared within a quantization group.

\textbf{PoT.} When $\hat{M_p} = \hat{M_n} = 1$, the dynamic range for positive and negative values can be adjusted by modifying the exponent. However, we observe that MXFP4's PoT frequently triggers max clamping in small group sizes, causing significant performance degradation. To address this, we propose an advanced PoT that mitigates max clamping by modifying the PoT decision rule (see Appendix~\ref{appendix:pot-r} for details). As shown in Fig.~\ref{fig:amxfp4}(a), the proposed PoT shared scale reduces LLaMA2 perplexity by approximately 4.

\textbf{FP8.} Although proposed PoT scale prevents clamping errors, its limited resolution still causes accuracy loss. To mitigate this issue, we propose using FP8 scales to leverage additional mantissa bits for finer rounding. However, as shown in Fig.~\ref{fig:amxfp4}(a), a 4-bit exponent results in a narrower dynamic range, which in turn increases perplexity compared to PoT. Therefore, we select FP8 with a 5-bit exponent ($E5M2$) as the shared scale, as these scales largely mitigate accuracy degradation caused by the limited resolution and narrower dynamic range (see Table~\ref{tab:shared_scale_ablation} for ablation studies).

\subsection{Asymmetric Microscaling Floating-Point}
\label{sec4.3}

Based on our exploration of the MX design space, we propose AMXFP4 (asymmetric microscaling 4-bit floating-point), which utilizes asymmetric FP8 shared scales. During multiplication, the shared scale is selected based on the signs of the two numbers. As shown in Fig.~\ref{fig:amxfp4}(b), this overhead remains minimal because the mantissa of the shared scale is only 2 bits, and the scale is computed once and shared within a group. To evaluate AMXFP4 on real hardware, we implement an AMXFP4 MAC unit via hardware synthesis by modifying the existing MX MAC unit~\cite{darvish2023shared}. Our evaluation shows that AMXFP4 incurs only about a 10\% overhead compared to MXFP4 (details are in Appendix~\ref{appendix:hw-analysis}).

%% file: Sections/experiments.tex
\section{Experiments}
\label{sec5}
In this section, we compare AMXFP4 with other formats and rotation-based methods. Unless otherwise specified, all experiments use the proposed FP8 shared scale across all formats (including INT4, MXFP4, and AMXFP4) for a fair comparison and quantize input operands for all decoder-layer matrix multiplications. Further details on quantization settings and benchmark descriptions are provided in Appendix~\ref{appendix:experimental_deatils}.

\subsection{Impact of Microscaling and Data Rotation}
\label{sec5.1}

\textbf{Microscaling vs. Data Rotation.}
We empirically validate the findings discussed in Sec.~\ref{sec3.2}, confirming that data rotation effectively mitigates activation outliers in configurations with large group sizes but has limited compatibility with microscaling. Table~\ref{tab:transformation} presents the impact of data rotation (randomized Hadamard transform) on Wikitext-2~\cite{merity2016pointerwiki} perplexity, with group sizes ranging from an entire row to 32.
When the group size spans an entire row, data rotation provides the best solution for MXFP4, outperforming asymmetric data representations. However, as the group size decreases, data rotation increases perplexity across all models with MXFP4, whereas AMXFP4 consistently reduces perplexity, achieving a 0.6-point reduction in LLaMA3-8B. This result further supports that outlier handling becomes less effective as group size decreases.

\textbf{INT4 vs. FP4.}
We extend our analysis to microscaling INT (MXINT) to assess whether the adverse effects of data rotation stem from FP’s non-uniform data representation. Similar to MXFP4, MXINT4 benefits from data rotation when the group size spans an entire row, significantly reducing perplexity compared to asymmetric representation (AMXINT4). However, at a group size of 32, data rotation tends to increase perplexity.
Notably, at group size 32, AMXINT4 achieves lower perplexity than MXFP4, but AMXFP4 achieves the lowest perplexity overall. This result demonstrates that our element format selection in Sec.~\ref{sec4.1} effectively enhances LLM accuracy.
\input{Tables/transformation-ppl}

\input{Tables/overfitting}
\textbf{Robustness to Calibration Set Distributions.}
Table~\ref{tab:overfitting} examines the sensitivity of QuaRot and SpinQuant to varying calibration set distributions. Perplexity is measured on PubMed~\cite{pubmeddataset} and Enron Emails~\cite{enron}, while accuracy is measured on PIQA~\cite{bisk2019piqa} and WinoGrande~\cite{sakaguchi2019winogrande}, using both matched and mismatched calibration/evaluation sets. QuaRot with GPTQ and SpinQuant substantially outperform the random Hadamard rotation but tend to show better accuracy on data observed during calibration. One exception is SpinQuant, which attains strong accuracy on both PIQA and WinoGrande when calibrated on PIQA, although results vary by about 2–3\% solely due to different calibration datasets. However, AMXFP4 remains unaffected by the calibration set and notably improves results and surpasses conventional calibration-based methods.

\begin{figure}[t]
\centering
\centerline{\includegraphics[width=0.95\columnwidth]{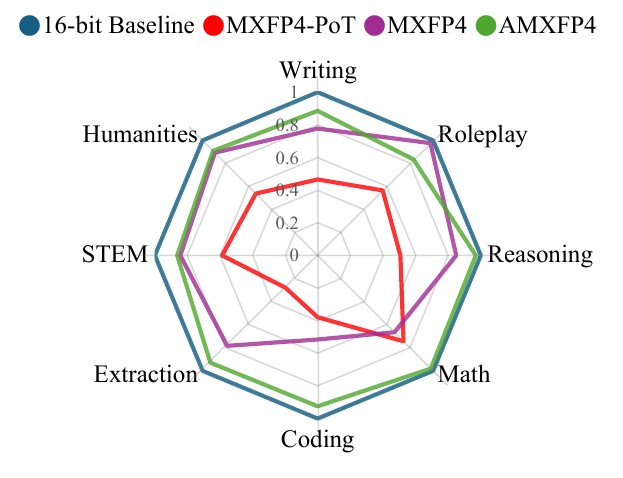}}
\caption{Normalized single score of MT-Bench (LLaMA2-Chat-7B). Absolute accuracies are in Table~\ref{tab:mx-mtbench} in Appendix.}
\label{fig:mt-bench}
\end{figure}

\subsection{Enhancing MX Performance}
\label{sec5.2}
In this section, we evaluate AMXFP4 against MXFP4 in practical applications, including chatbots, visual tasks, and long-document question answering. To assess our improvements over the \textit{MX-compliant} format, we also include MXFP4 with PoT shared scale (MXFP4-PoT) from Sec~\ref{sec4.2} as a baseline for comparison. We find that the superior performance of AMXFP4 over MXFP4 is consistently observed across various architectures and scales, including language modeling tasks, encoder-decoder models, and 70B-scale LLMs. These extended results are provided in Appendix~\ref{appendix:ablation}.

\textbf{Multi-Turn Chatbot Tasks.}
Quantization adversely affects the conversational capabilities of chatbots~\cite{lee-etal-2024-improving-conversational}; therefore, we conduct an MT-Bench evaluation~\cite{zheng2023judging} on LLaMA2-Chat-7B~\cite{touvron2023llama2}. Fig.~\ref{fig:mt-bench} presents the normalized scores with the 16-bit baseline score set to 1. While MXFP4 inference shows severe performance degradation across all categories, AMXFP4 demonstrates recovery of conversational abilities close to the baseline. Fig.~\ref{fig:chat-example} and ~\ref{fig:mt-bench-example} provide detailed examples, showing that while MXFP4 generates unhelpful sentences, AMXFP4 produces responses that are genuinely helpful.

\input{Tables/mx-lmm-eval}
\textbf{Visual Tasks.} Table~\ref{tab:mx-lmm-eval} presents results on four multi-modal benchmarks~\cite{zhang2024lmmsevalrealitycheckevaluation} using LLaVA1.6-7B~\cite{liu2023llava}. AMXFP4 improves MXFP4 scores by approximately 3.3 points on benchmarks such as ChartQA~\cite{masry-etal-2022-chartqa}, highlighting the significant advantages of asymmetric data representation in VLMs (example is shown in Fig.~\ref{fig:chartqa}).

\begin{figure}[t]
\centering
\centerline{\includegraphics[width=\columnwidth]{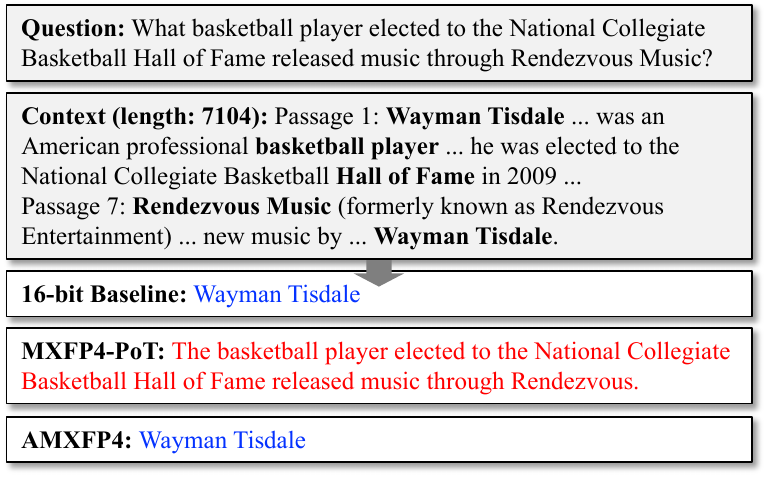}}
\caption{LongBench-E results on LLaMA2-Chat-7B.}
\label{fig:longbench}
\end{figure}

\textbf{Long-Context Tasks.} We conduct the LongBench-E~\cite{bai-etal-2024-longbench} evaluation to assess the effectiveness of AMXFP4 in long-context scenarios. As shown in Fig.~\ref{fig:longbench}, while MXFP4-PoT’s generation quality significantly degrades on questions with lengthy contexts, AMXFP4 produces answers identical to the baseline.
Detailed scores across 13 benchmarks, categorized by context length, are presented in Table~\ref{tab:longbench}. The results indicate that AMXFP4 outperforms MXFP4, achieving over a 2\% accuracy improvement for context lengths exceeding 8K.

\subsection{Comparison with Commercial MXFP4}
\label{sec5.3}

Recently, NVFP4~\cite{tensorrt-llm} adopts a smaller group size of 16 and employs a double-scaling strategy, which combines a tensor-wise FP32 shared scale with a group-wise FP8 (E4M3) shared scale. We evaluate whether our proposed asymmetric shared scale enhances the recently deplyed commercial MXFP4 by evaluating ANVFP4 (Asymmetric NVFP4) on Common-Sense Question Answering (CSQA)~\cite{talmor2019commonsenseqa} and MMLU~\cite{mmlu} benchmarks. As shown in Table~\ref{tab:nvfp4}, when GS=32, AMXFP4 and ANVFP4 surpass NVFP4 in accuracy, indicating that the asymmetric data representation offers a greater improvement than double scaling strategy. Notably, in the NVFP4 setting with GS=16, ANVFP4 increases MMLU accuracy by about 3\%, which aligns with our observation that asymmetry becomes more beneficial at smaller group sizes.

\begin{table}[t]
\centering
\resizebox{\columnwidth}{!}{%
\begin{tabular}{cc|ccc|ccc}
\hline
\multicolumn{1}{c|}{} &  & \multicolumn{3}{c|}{MMLU Accuracy (\%) ↑} & \multicolumn{3}{c}{CSQA Accuracy (\%) ↑} \\
\multicolumn{1}{c|}{\multirow{-2}{*}{GS}} & \multirow{-2}{*}{Data Format} & 2-7B & 2-13B & 3-8B & 2-7B & 2-13B & 3-8B \\ \hline
\multicolumn{2}{c|}{16-bit Baseline} & 41.3 & 50.5 & 62.0 & 64.9 & 67.3 & 69.2 \\ \hline
\multicolumn{1}{c|}{} & MXFP4-PoT & 29.2 & 37.9 & 43.1 & 59.4 & 62.2 & 58.6 \\
\multicolumn{1}{c|}{} & MXFP4 & 33.6 & 42.8 & 49.5 & 61.6 & 65.1 & 62.0 \\
\multicolumn{1}{c|}{} & \cellcolor[HTML]{ECF4FF}AMXFP4 & \cellcolor[HTML]{ECF4FF}\textbf{36.3} & \cellcolor[HTML]{ECF4FF}{\ul 45.0} & \cellcolor[HTML]{ECF4FF}{\ul 52.8} & \cellcolor[HTML]{ECF4FF}{\ul 62.0} & \cellcolor[HTML]{ECF4FF}{\ul 64.9} & \cellcolor[HTML]{ECF4FF}{\ul 62.2} \\
\multicolumn{1}{c|}{} & NVFP4 & 32.9 & 44.5 & 51.9 & 61.4 & \textbf{65.0} & 61.9 \\
\multicolumn{1}{c|}{\multirow{-5}{*}{32}} & \cellcolor[HTML]{ECF4FF}ANVFP4 & \cellcolor[HTML]{ECF4FF}{\ul 34.8} & \cellcolor[HTML]{ECF4FF}\textbf{45.8} & \cellcolor[HTML]{ECF4FF}\textbf{54.0} & \cellcolor[HTML]{ECF4FF}\textbf{62.2} & \cellcolor[HTML]{ECF4FF}64.7 & \cellcolor[HTML]{ECF4FF}\textbf{62.9} \\ \hline
\multicolumn{1}{c|}{} & NVFP4 & {\ul 34.0} & {\ul 45.9} & {\ul 54.6} & \textbf{62.6} & {\ul 65.3} & {\ul 63.4} \\
\multicolumn{1}{c|}{\multirow{-2}{*}{16}} & \cellcolor[HTML]{ECF4FF}{\color[HTML]{000000} ANVFP4} & \cellcolor[HTML]{ECF4FF}{\color[HTML]{000000} \textbf{37.3}} & \cellcolor[HTML]{ECF4FF}{\color[HTML]{000000} \textbf{47.7}} & \cellcolor[HTML]{ECF4FF}{\color[HTML]{000000} \textbf{57.1}} & \cellcolor[HTML]{ECF4FF}{\color[HTML]{000000} {\ul 62.2}} & \cellcolor[HTML]{ECF4FF}{\color[HTML]{000000} \textbf{66.2}} & \cellcolor[HTML]{ECF4FF}{\color[HTML]{000000} \textbf{64.9}} \\ \hline
\end{tabular}%
}
\caption{MMLU and CSQA results on LLaMA models.}
\label{tab:nvfp4}
\end{table}

\subsection{Ablation Studies}
\label{sec5.4}

\begin{table}[t]
\centering
\resizebox{0.9\columnwidth}{!}{%
\begin{tabular}{c|cc|cc}
\hline
Group Size & \multicolumn{2}{c|}{Row} & \multicolumn{2}{c}{32} \\ \hline
Rotation & MXFP4 & AMXFP4 & MXFP4 & AMXFP4 \\ \hline
- & 97.60 & 28.99 & 5.93 & \textbf{5.85} \\
Random & 10.78 & 11.76 & 9.23 & 8.02 \\
SpinQuant & 6.37 & \textbf{6.33} & 6.10 & 6.04 \\ \hline
\end{tabular}%
}
\caption{Perplexity on Wikitext-2 under different rotation types and group sizes (LLaMA2-7B).}
\label{tab:spinquant}
\end{table}

\textbf{Extension to SpinQuant.}
To validate the generality of our findings in Sec.~\ref{sec5.1}, we extend the rotation experiments to SpinQuant~\cite{liu2024spinquant}. As shown in Table~\ref{tab:spinquant}, SpinQuant achieves lower perplexity at row-level granularity but fails to outperform the no-rotation baseline at group size 32, consistent with our earlier observations. Moreover, it exhibits overfitting to the calibration set (Table~\ref{tab:overfitting}), whereas AMXFP4 remains effective without calibration. These results confirm that the destructive interaction between rotation and microscaling persists even with learned rotation strategies.

\begin{table}[t]
\centering
\resizebox{0.9\columnwidth}{!}{%
\begin{tabular}{c|c|c|cc}
\hline
Method & Data Format & PPL ↓ & Memory & Time \\ \hline
Direct-Cast & 16-bit Baseline & 6.14 & - & - \\ \hline
Direct-Cast & MXFP4-PoT & 7.70 & - & - \\
\rowcolor[HTML]{ECF4FF} 
Direct-Cast & AMXFP4 & \textbf{6.97} & - & - \\ \hline
QAT & MXFP4-PoT & 6.68 & 148GB & 4h 30m \\
\rowcolor[HTML]{ECF4FF} 
QAT & AMXFP4 & \textbf{6.33} & 148GB & 4h 30m \\ \hline
\end{tabular}%
}
\caption{QAT results on LLaMA3-8B using the Wikitext-2 dataset. ``Memory'' and ``Time'' refer to the GPU memory usage and fine-tuning time required for QAT, measured on two A100-80GB GPUs.}
\label{tab:qat}
\end{table}

\textbf{Quantization-Aware Training (QAT).}
We investigate whether QAT can mitigate the perplexity gap between MXFP4 and AMXFP4. As shown in Table~\ref{tab:qat}, under direct-cast inference without calibration, MXFP4 incurs a perplexity increase of 1.6 compared to the 16-bit baseline on LLaMA3-8B, while AMXFP4 shows a smaller degradation of only 0.8. Applying QAT significantly improves performance for both formats, with AMXFP4 achieving perplexity nearly on par with the baseline, and consistently outperforming MXFP4 even after training. However, it is worth noting that QAT introduces substantial computational overhead, requiring approximately 150GB of GPU memory and 5 hours of fine-tuning time, along with additional cost for hyperparameter tuning. Full training configurations are detailed in Appendix~\ref{appendix:experimental_deatils}.

\input{Tables/w3a3}
\textbf{More Aggressive Quantization.}
We compare QuaRot and AMXFP under a 3-bit setting (W3A3) in Table~\ref{tab:w3a3}. While QuaRot with GPTQ suffers a severe degradation exceeding 30 in W3A3, AMXFP3 achieves a perplexity degradation of only 1.7 in direct-cast inference, highlighting AMXFP’s potential in lower-precision settings.

\textbf{Attention-Only Quantization.}
AMXFP4 is designed for full-model quantization using unified low-precision formats, but we also conduct a scope-aligned experiment by restricting its application to attention components only (\textit{Query}, \textit{Key}, \textit{Self-attention map} and \textit{Value}) to match the selective quantization setting adopted in SageAttention~\cite{zhang2025sageattention}. As shown in Table~\ref{tab:attention-eval}, AMXFP4 achieves comparable accuracy to SageAttention despite using more aggressive 4-bit quantization with unified FP8 scaling, and exhibits slightly improved performance when the shared scale precision is increased to FP16.

Recent attention-only quantization methods have extended this line of work to lower-bit formats including INT4~\cite{zhang2024sageattention2} and NVFP4~\cite{zhang2025sageattention3}, in response to hardware precision scaling trends. Our results suggest that even in such selective quantization settings, explicit handling of asymmetry—as enabled by AMXFP4—can offer meaningful advantages.

\begin{table}[t]
\centering
\resizebox{\columnwidth}{!}{%
\begin{tabular}{c c c c c}
\hline
Method & \makecell{Shared\\Scale} & \makecell{Q/K\\Format} & \makecell{SA/V\\Format} & \makecell{Wiki2 $\downarrow$\\ / MMLU $\uparrow$} \\
\hline
SageAttention & FP32 & INT8   & FP16   & 5.47 / 38.38 \\
MXFP4         & FP8  & MXFP4  & MXFP4  & 5.91 / 37.13 \\
\rowcolor[HTML]{ECF4FF} 
AMXFP4        & FP8  & AMXFP4 & AMXFP4 & 5.81 / 38.26 \\
\rowcolor[HTML]{ECF4FF} 
AMXFP4        & FP16 & AMXFP4 & AMXFP4 & 5.69 / 39.53 \\
\hline
\end{tabular}
}
\caption{Evaluation of attention-only quantization using AMXFP4 compared to SageAttention (LLaMA2-7B).}
\label{tab:attention-eval}
\end{table}

\begin{table}[t]
\centering
\resizebox{0.9\columnwidth}{!}{%
\begin{tabular}{c|ccc}
\hline
Data Format & Area-Memory & Power-Area & \begin{tabular}[c]{@{}c@{}}Power-Area\\ -Memory\end{tabular} \\ \hline
FP16 & 1.00$\times$ & 1.00$\times$ & 1.00$\times$ \\ \hline
MXFP4-PoT & 10.44$\times$ & 7.62$\times$ & 28.67$\times$ \\
MXFP4 & 9.23$\times$ & 5.65$\times$ & 21.41$\times$ \\
\rowcolor[HTML]{ECF4FF} 
AMXFP4 & 8.32$\times$ & 4.58$\times$ & 16.50$\times$ \\ \hline
\end{tabular}%
}
\caption{Hardware comparison between MXFP4 and AMXFP4.}
\label{tab:hw-analysis}
\end{table}

\subsection{Hardware Evaluation for AMXFP4}
\label{appendix:hw-analysis}

To evaluate the hardware efficiency of AMXFP4, we follow and extend the analysis methodology of~\cite{darvish2023shared}, focusing on area and memory cost. We implement a fully custom MX-compatible MAC unit and its AMX extension, and synthesize both using Synopsys Design Compiler under a 4nm CMOS process (0.675V, 1.1GHz). The group-wise representation of MX decouples dot products from scaling operations, enabling efficient MAC design with minimal inter-group overhead (Fig.~\ref{fig:mx-mac}). As shown in Table~\ref{tab:hw-analysis}, our MX-compatible MAC reduces area and memory usage by over 8$\times$, consistent with recent accelerator designs adopting MXFP4~\cite{b100,maia}. AMXFP4 introduces sign-aware mantissa scaling for asymmetric group scales, yet adds only 10\% overhead due to the narrow mantissa width and scale reuse within groups.

%% file: Tables/transformation-ppl.tex
\begin{table}[t]
\centering
\resizebox{\columnwidth}{!}{%
\begin{tabular}{ccc|ccc}
\hline
\multicolumn{1}{c|}{} & \multicolumn{1}{c|}{} &  & \multicolumn{3}{c}{LLaMA} \\
\multicolumn{1}{c|}{\multirow{-2}{*}{\begin{tabular}[c]{@{}c@{}}Group\\ Size\end{tabular}}} & \multicolumn{1}{c|}{\multirow{-2}{*}{\begin{tabular}[c]{@{}c@{}}Data\\ Rotation\end{tabular}}} & \multirow{-2}{*}{\begin{tabular}[c]{@{}c@{}}Data\\ Format\end{tabular}} & 2-7B & 2-13B & 3-8B \\ \hline
\multicolumn{3}{c|}{FP16 Baseline} & 5.47 & 4.88 & 6.14 \\ \hline
\multicolumn{1}{c|}{} & \multicolumn{1}{c|}{} & MXINT4 & NaN & 2988.82 & 2603.42 \\
\multicolumn{1}{c|}{} & \multicolumn{1}{c|}{} & AMXINT4 & 2045.70 & 364.96 & 1800.44 \\
% \multicolumn{1}{c|}{} & \multicolumn{1}{c|}{} & MXFP4-PoT & 1850.70 & 480.80 & 412.10 \\
\multicolumn{1}{c|}{} & \multicolumn{1}{c|}{} & MXFP4 & 475.62 & 99.33 & 85.07 \\
\multicolumn{1}{c|}{} & \multicolumn{1}{c|}{\multirow{-4}{*}{-}} & AMXFP4 & 44.75 & 33.79 & 40.33 \\ \cline{2-6} 
\multicolumn{1}{c|}{} & \multicolumn{1}{c|}{} & MXINT4 & 47.55 & 35.32 & 100.95 \\
\multicolumn{1}{c|}{} & \multicolumn{1}{c|}{} & AMXINT4 & 16.60 & 13.94 & 35.90 \\
% \multicolumn{1}{c|}{} & \multicolumn{1}{c|}{} & MXFP4-PoT & 28.22 & 15.19 & 48.74 \\
\multicolumn{1}{c|}{} & \multicolumn{1}{c|}{} & MXFP4 & \textbf{11.88} & \textbf{10.81} & 13.27 \\
\multicolumn{1}{c|}{\multirow{-8}{*}{Row}} & \multicolumn{1}{c|}{\multirow{-4}{*}{\checkmark}} & AMXFP4 & 12.05 & 11.54 & \textbf{12.13} \\ \hline
\multicolumn{1}{c|}{} & \multicolumn{1}{c|}{} & MXINT4 & 7.01 & 6.11 & 9.01 \\
\multicolumn{1}{c|}{} & \multicolumn{1}{c|}{} & AMXINT4 & 6.33 & 5.55 & 9.62 \\
% \multicolumn{1}{c|}{} & \multicolumn{1}{c|}{} & MXFP4-PoT & 7.83 & 6.98 & 11.17 \\
\multicolumn{1}{c|}{} & \multicolumn{1}{c|}{} & MXFP4 & 6.49 & 5.69 & 8.35 \\
\multicolumn{1}{c|}{} & \multicolumn{1}{c|}{\multirow{-4}{*}{-}} & \cellcolor[HTML]{ECF4FF}AMXFP4 & \cellcolor[HTML]{ECF4FF}\textbf{6.22} & \cellcolor[HTML]{ECF4FF}\textbf{5.47} & \cellcolor[HTML]{ECF4FF}\textbf{7.72} \\ \cline{2-6} 
\multicolumn{1}{c|}{} & \multicolumn{1}{c|}{} & MXINT4 & 7.90 & 6.18 & 9.96 \\
\multicolumn{1}{c|}{} & \multicolumn{1}{c|}{} & AMXINT4 & 6.75 & 5.75 & 8.25 \\
% \multicolumn{1}{c|}{} & \multicolumn{1}{c|}{} & MXFP4-PoT & 14.55 & 7.93 & 13.32 \\
\multicolumn{1}{c|}{} & \multicolumn{1}{c|}{} & MXFP4 & 10.09 & 6.89 & 9.48 \\
\multicolumn{1}{c|}{\multirow{-8}{*}{MX (32)}} & \multicolumn{1}{c|}{\multirow{-4}{*}{\checkmark}} & AMXFP4 & 8.36 & 6.35 & 9.20 \\ \hline
\end{tabular}%
}
\caption{Wikitext-2 perplexity results by group size with and without data rotation applied (lower is better).}
\label{tab:transformation}
\end{table}

%% file: Tables/overfitting.tex
\begin{table}[t]
\centering
\resizebox{\columnwidth}{!}{%
\begin{tabular}{cc|c|cc|cc|
>{\columncolor[HTML]{ECF4FF}}c }
\hline
\multicolumn{1}{c|}{LLaMA} & \begin{tabular}[c]{@{}c@{}}Eval\\ Dataset\end{tabular} & QuaRot & \multicolumn{2}{c|}{\begin{tabular}[c]{@{}c@{}}QuaRot\\ + GPTQ\end{tabular}} & \multicolumn{2}{c|}{SpinQuant} & AMXFP4 \\ \hline
\multicolumn{2}{c|}{Calib Dataset} & - & PM & EE & PM & EE & - \\ \hline
\multicolumn{1}{c|}{} & PM ↓ & 7.7 & {\color[HTML]{156082} 5.4} & {\color[HTML]{C00000} 5.5} & {\color[HTML]{156082} 5.7} & {\color[HTML]{C00000} 5.9} & \textbf{5.3} \\
\multicolumn{1}{c|}{\multirow{-2}{*}{2-7B}} & EE ↓ & 7.9 & {\color[HTML]{C00000} 6.3} & {\color[HTML]{156082} {6.2}} & {\color[HTML]{C00000} 6.8} & {\color[HTML]{156082} 6.3} & \textbf{6.1} \\ \hline
\multicolumn{1}{c|}{} & PM ↓ & 9.4 & {\color[HTML]{156082} 7.4} & {\color[HTML]{C00000} 7.6} & {\color[HTML]{156082} 7.5} & {\color[HTML]{C00000} 7.7} & \textbf{6.8} \\
\multicolumn{1}{c|}{\multirow{-2}{*}{3-8B}} & EE ↓ & 12.9 & {\color[HTML]{C00000} 10.7} & {\color[HTML]{156082} 10.2} & {\color[HTML]{C00000} 10.7} & {\color[HTML]{156082} 10.0} & \textbf{9.4} \\ \hline
\multicolumn{2}{c|}{Calibration Dataset} & - & PQ & WG & PQ & WG & - \\ \hline
\multicolumn{1}{c|}{} & PQ ↑ & 72.0 & {\color[HTML]{156082} 77.4} & {\color[HTML]{C00000} 76.2} & {\color[HTML]{156082} 76.4} & {\color[HTML]{C00000} 73.1} & \textbf{77.8} \\
\multicolumn{1}{c|}{\multirow{-2}{*}{2-7B}} & WG ↑ & 60.1 & {\color[HTML]{C00000} 65.3} & {\color[HTML]{156082} 65.9} & {\color[HTML]{156082} 66.4} & {\color[HTML]{C00000} 64.0} & \textbf{67.5} \\ \hline
\end{tabular}%
}
\vspace{0.5em}
\noindent\scriptsize\raggedright PM: PubMed, EE: Enron Emails, PQ: PIQA, WG: WinoGrande
\vspace{-3mm}
\caption{Impact of overfitting: Calibration on different data distribution on LLaMA models.}
\label{tab:overfitting}
\end{table}

%% file: Tables/mx-lmm-eval.tex
% \usepackage{graphicx}
\begin{table}[t]
\centering
\resizebox{\columnwidth}{!}{%
\begin{tabular}{c|cccc}
\hline
% Data Format & VQA-T & DocVQA & OCRBench & ChartQA \\ \hline
% 16-bit Baseline & 64.84 & 74.46 & 52.40 & 54.72 \\ \hline
% MXFP4-PoT & 50.05 & 52.85 & 33.70 & 36.76 \\
% MXFP4-FP8 & 57.88 & 64.26 & 43.40 & 46.20 \\ \hline
% AMXFP4-PoT & 53.19 & 56.55 & 37.20 & 39.80 \\
Data Format & VQA-T & DocVQA & OCRBench & ChartQA \\ \hline
16-bit Baseline & 64.84 & 74.46 & 52.40 & 54.72 \\ \hline
MXFP4-PoT & 50.05 & 52.85 & 33.70 & 36.76 \\
MXFP4 & 57.88 & 64.26 & 43.40 & 46.20 \\
\rowcolor[HTML]{ECF4FF} 
AMXFP4 & \textbf{59.13} & \textbf{66.98} & \textbf{43.90} & \textbf{49.48} \\ \hline
\end{tabular}%
}
\caption{LLaVA1.6-7B inference results on multi-modal visual question-answering benchmarks.}
\label{tab:mx-lmm-eval}
\end{table}

%% file: Tables/w3a3.tex
% \usepackage{multirow}
% \usepackage{graphicx}
\begin{table}[t]
\centering
\resizebox{0.9\columnwidth}{!}{%
\begin{tabular}{cc|cc}
\hline
\multicolumn{1}{c|}{Method} & Format & LLaMA2-7B & LLaMA2-13B \\ \hline
\multicolumn{2}{c|}{16-bit Baseline} & 5.47 & 4.88 \\ \hline
\multicolumn{1}{c|}{QuaRot-RTN} &  & 1032.30 & 1105.95 \\
\multicolumn{1}{c|}{QuaRot-GPTQ} & \multirow{-2}{*}{INT} & 38.47 & 37.42 \\ \hline
% \multicolumn{1}{c|}{MXFP3-PoT} &  & NaN & 39.79 \\
\multicolumn{1}{c|}{\cellcolor[HTML]{ECF4FF}AMXFP3} & \cellcolor[HTML]{ECF4FF}MX & \cellcolor[HTML]{ECF4FF}\textbf{8.40} & \cellcolor[HTML]{ECF4FF}\textbf{6.53} \\ \hline
\end{tabular}%
}
\caption{Wikitext-2 perplexity results on 3-bit inference.}
\label{tab:w3a3}
\end{table}

%% file: Sections/conclusion.tex
\section{Conclusion}
To meet the computational demands of large language models (LLMs) with extended contexts, we introduce Asymmetric Microscaling 4-bit Floating-Point (AMXFP4), which uses asymmetric shared scales to handle outliers and quantization asymmetry. AMXFP4 provides direct 4-bit inference with high accuracy, outperforming MXFP4 and other techniques for efficient, calibration-free inference.

\section*{Acknowledgement}
This work was supported by Institute of Information \& communications Technology Planning \& Evaluation (IITP) (under the artificial intelligence semiconductor support program to nurture the best talents, IITP-2025-RS-2023-00253914, and No.RS-2025-02214497, Development of low-level optimization program API technology for AI semiconductors) and National Research Foundation of Korea (NRF) (No. RS-2025-00561961) grant funded by the Korea government (MSIT).

%% file: Sections/appendix-mx.tex
\clearpage

\section{Comparison with Rotation Techniques}
\label{appendix:rotation}
Rotation-based methods, such as QuaRot and SpinQuant, typically avoid quantizing query and softmax output, and require on additional calibration, which introduces the following drawbacks:

\begin{figure}[t]
\centering
\centerline{\includegraphics[width=\columnwidth]{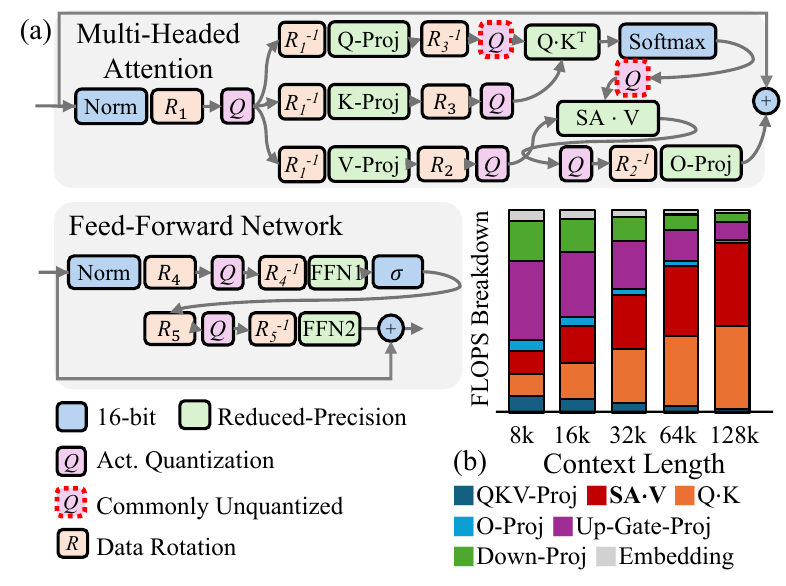}}
\caption{(a) Illustration of where reduced-precision matrix multiplication and data transformation are applied within a Transformer decoder layer. QuaRot and SpinQuant do not quantize the Query and Softmax outputs (red dotted box). (b) FLOPS breakdown of LLaMA3-8B in the prefill stage based on context length.}
\label{fig:rotation}
\end{figure}

\textbf{High-Precision Query and Softmax Output.}
Fig.~\ref{fig:rotation}(a) illustrates how rotation-based methods apply rotation and quantization in reduced-precision LLM inference. While these techniques make activations more quantization-friendly, they do not quantize the softmax output. As shown in Fig.~\ref{fig:rotation}(b), as context length increases, the dominant FLOPS in the prefill stage come from query-key multiplication and attention operations, including softmax output (self-attention map; SA) and value multiplication. Processing these operations in high precision undermines the benefits of reduced-precision inference, limiting overall efficiency.

\input{Tables/calibration-csqa}
\textbf{Calibration Overhead.}
Table~\ref{tab:calibration-csqa} displays the effects of varying calibration settings (dataset, sequence length, and number of samples) on Wikitext-2 perplexity, ARC-Challenge~\cite{allenai:arc} and WinoGrande accuracy for QuaRot and SpinQuant. When using QuaRot alone, CSQA accuracy drops by 10\%. When combined QuaRot with GPTQ, results depend on calibration settings; using only 32 calibration samples leads to a 2.4\% reduction in WinoGrande accuracy compared to using 128 samples. SpinQuant, which trains a rotation matrix, achieves higher accuracy than QuaRot alone but increases calibration time by approximately 6$\times$ and exhibits greater sensitivity to the calibration set. When calibrated with the PTB~\cite{marcus-etal-1993-building-ptb} dataset instead of Wikitext-2, perplexity on Wikitext-2 rises by around 0.9. Our proposed AMXFP4 shows minimal performance degradation compared to the baseline and remains unaffected by calibration settings.

\section{MX Format Details and Emulation Framework}
\label{appendix:mx}

\subsection{MX Configuration}
\label{secA.1}

\begin{algorithm}
\small
\label{alg:mx}
\begin{algorithmic}[1]
\STATE Quantize vector elements ($\{V_i\}_{i=1}^k$) into MX format
\STATE 
$shared\_exp \gets \lfloor \log_2(\max_{i}(|V_i|)) \rfloor - emax_{elem}$
\STATE $X \gets 2^{shared\_exp}$
\FOR{$i=1$ to $k$}
  \STATE $P_i = \text{quantize}(V_i / X)$, clamping normal numbers
\ENDFOR
\STATE\textbf{return} $X,\>\{P_i\}_{i=1}^k$
\end{algorithmic}
\caption{Quantization procedure in MX format. Algorithm is adapted from~\cite{rouhani2023microscaling}.}
\end{algorithm}

\input{Tables/mx_configs}

As the MX format is our primary focus for improvement, we aim to provide detailed information on it. We follow the MX format configuration and quantization procedure as~\cite{ocp,rouhani2023microscaling}. The MX format offers a variety of bit-configurations for elements, ranging from 8 bits to 4 bits, while specifying only an 8-bit PoT for the shared scale. The process to determine this 8-bit PoT follows an Algorithm 1. As described in the entire quantization procedure, MX considers the maximum data value to determine the shared scale, performing a floor operation after extracting the exponent of the element’s maximum value with log2.

\begin{figure}[t]
\centerline{\includegraphics[width=\columnwidth]{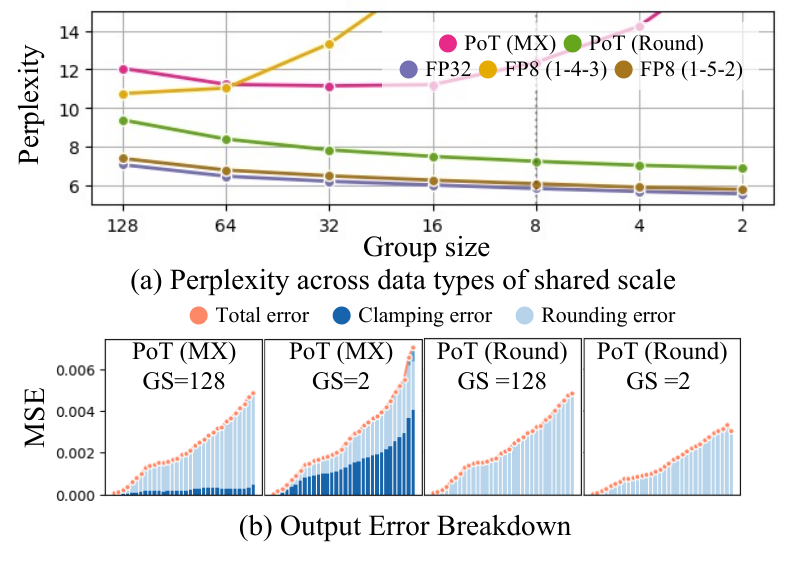}}
\caption{Impact of shared scale (LLaMA2-7B). More results on other models and data formats are in Table~\ref{tab:shared_scale_ablation}.}
\label{fig:shared_scale}
\end{figure}

\subsection{Determining PoT Shared Scale: Floor vs. Round}
\label{appendix:pot-r}
As illustrated in Fig.~\ref{fig:shared_scale}(a), an undesirable performance degradation occurs in PoT scales as group size decreases. To analyze this degradation, we decompose the output error into maximum clamping error and rounding error. As shown in Fig.~\ref{fig:shared_scale}(b), with a group size of 2, the rounding error reduces significantly, while the maximum clamping error increases sharply, resulting in a net error rise. This issue is attributed to the floor operation on the exponent in MX, which introduces clamping error. To overcome maximum clamping errors while maintaining the hardware efficiency of PoT shared scales, we replace flooring with rounding. This exponent rounding approach significantly lowers total error, enhancing performance, as demonstrated in Fig.~\ref{fig:shared_scale}(a) and (b).

\begin{figure}[t]
\centering
\centerline{\includegraphics[width=\columnwidth]{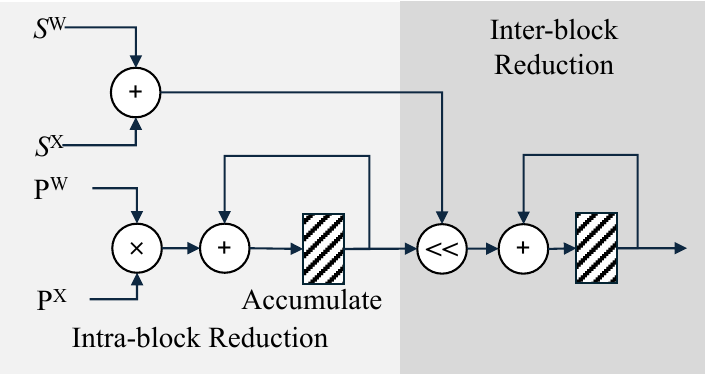}}
\caption{MX dot-product architecture.}
\label{fig:mx-mac}
\end{figure}

\subsection{Code Snippet of Our Framework}
\label{appendix:emulation-library}
  As shown in the below example, our proposed AMXFP4 applies different shared scales to positive and negative numbers, enabling more refined value representation compared to MXFP4. Additionally, the PoT shared scale significantly clamps the largest value in the input, 31, to 24, while the FP8 shared scale, using the same number of bits, more precisely quantizes 31 to 30.

\begin{python}
class MXQuantizer(object):
    def __init__(self, elem_format, group_size, scale_mode):
        self.elem_format = elem_format # Element Format
        self.group_size = group_size # group Size
        self.scale_mode = scale_mode # Shared Scale Type
        self.mx_specs = MxSpecs(
            a_elem_format=self.elem_format,
            group_size=self.group_size,
            custom_cuda=True,
            scale_mode=scale_mode,
        )
    def quantize(self, x):
        qx = quantize_mx_op(
            x,
            self.mx_specs,
            elem_format=self.elem_format,
            axes=[-1],
        )
        return qx

# Example: Asymmetrically distributed tensor with a single row
x = torch.linspace(-4.9, 31, 1024)

# MXFP4
mx_fp4 = MXQuantizer(elem_format='fp4_e2m1', group_size=-1, scale_mode=0)
qx_mx_fp4 = mx_fp4.quantize(x)
# AMXFP4 (Shared Scale: PoT)
mx_fp4_asym = MXQuantizer(elem_format='fp4_e2m1_asym', group_size=-1, scale_mode=0)
qx_mx_fp4_asym = mx_fp4_asym.quantize(x)
# AMXFP4 (Shared Scale: FP8)
mx_fp4_asym_fp8scale = MXQuantizer(elem_format='fp4_e2m1_asym', group_size=-1, scale_mode=152)
qx_mx_fp4_asym_fp8scale = mx_fp4_asym_fp8scale.quantize(x)

# Quantized tensor
print(qx_mx_fp4.unique()) # MXFP4
>> tensor([-4., -2.,  0.,  2.,  4.,  6.,  8., 12., 16., 24.], device='cuda:0')
print(qx_mx_fp4_asym.unique()) # AMXFP4 (Shared Scale: PoT)
>> tensor([-4.0000, -3.0000, -2.0000, -1.5000, -1.0000, -0.5000,  0.0000,  2.0000,
         4.0000,  6.0000,  8.0000, 12.0000, 16.0000, 24.0000], device='cuda:0')
print(qx_mx_fp4_asym_fp8scale.unique()) # AMXFP4 (Shared Scale: FP8)
>> tensor([-5.2500, -3.5000, -2.6250, -1.7500, -1.3125, -0.8750, -0.4375,  0.0000,
         2.5000,  5.0000,  7.5000, 10.0000, 15.0000, 20.0000, 30.0000],
       device='cuda:0')
\end{python}

%% file: Tables/calibration-csqa.tex
\begin{table}[t]
\centering
\resizebox{\columnwidth}{!}{%
\begin{tabular}{ccc|c|cc}
\hline
\multicolumn{1}{c|}{} & \multicolumn{1}{c|}{} &  & PPL↓ & \multicolumn{2}{c}{Accuracy↑} \\ \cline{4-6} 
\multicolumn{1}{c|}{\multirow{-2}{*}{Rotation}} & \multicolumn{1}{c|}{\multirow{-2}{*}{\begin{tabular}[c]{@{}c@{}}Calibset-\\ SeqLen-Samples\end{tabular}}} & \multirow{-2}{*}{\begin{tabular}[c]{@{}c@{}}Calib. Time\\ (A100)\end{tabular}} & Wiki & ARC-C & WG \\ \hline
\multicolumn{3}{c|}{16-bit Baseline} & 5.47 & 46.33 & 69.30 \\ \hline
\multicolumn{1}{c|}{QuaRot} & \multicolumn{1}{c|}{-} & - & 8.38 & 36.26 & 60.06 \\ \hline
\multicolumn{1}{c|}{} & \multicolumn{1}{c|}{Wiki-2048-128} &  & 6.08 & 41.64 & 66.22 \\
\multicolumn{1}{c|}{} & \multicolumn{1}{c|}{Wiki-1024-128} &  & 6.06 & 42.32 & 65.59 \\
\multicolumn{1}{c|}{} & \multicolumn{1}{c|}{Wiki-2048-64} &  & 6.11 & 41.64 & 65.51 \\
\multicolumn{1}{c|}{} & \multicolumn{1}{c|}{Wiki-2048-32} &  & 6.11 & 41.55 & 63.85 \\
\multicolumn{1}{c|}{} & \multicolumn{1}{c|}{PTB-2048-128} &  & 6.16 & 42.15 & 65.43 \\
\multicolumn{1}{c|}{\multirow{-6}{*}{\begin{tabular}[c]{@{}c@{}}QuaRot+\\ GPTQ\end{tabular}}} & \multicolumn{1}{c|}{PTB-1024-128} & \multirow{-6}{*}{$\sim$20 min} & 6.12 & 41.72 & 66.54 \\ \hline
\multicolumn{1}{c|}{} & \multicolumn{1}{c|}{Wiki-2048-100} &  & 6.25 & 38.65 & 64.72 \\
\multicolumn{1}{c|}{} & \multicolumn{1}{c|}{Wiki-1024-100} &  & 6.32 & 40.87 & 63.77 \\
\multicolumn{1}{c|}{} & \multicolumn{1}{c|}{PTB-2048-100} &  & 7.11 & 38.74 & 60.30 \\
\multicolumn{1}{c|}{\multirow{-4}{*}{SpinQuant}} & \multicolumn{1}{c|}{PTB-1024-100} & \multirow{-4}{*}{$\sim$2 hours} & 7.14 & 37.71 & 63.54 \\ \hline
\rowcolor[HTML]{ECF4FF} 
\multicolumn{3}{c|}{\cellcolor[HTML]{ECF4FF}AMXFP4 (direct-cast, no calibration)} & \textbf{5.93} & \textbf{42.83} & \textbf{67.32} \\ \hline
\end{tabular}%
}
\caption{Calibration overhead on LLaMA2-7B.}
\label{tab:calibration-csqa}
\end{table}

%% file: Tables/mx_configs.tex
% \usepackage{multirow}
% \usepackage{graphicx}
\begin{table}[t]
\centering
\resizebox{\columnwidth}{!}{%
\begin{tabular}{c|c|c|c|c}
\hline
Name & \begin{tabular}[c]{@{}c@{}}Element Data\\ Type\end{tabular} & Element Bits & Group Size & Shared Scale \\ \hline
\multirow{2}{*}{MXFP8} & FP8 (E5M2) & \multirow{2}{*}{8} & \multirow{6}{*}{32} & \multirow{6}{*}{8-bit PoT} \\
 & FP8 (E4M3) &  &  &  \\ \cline{1-3}
\multirow{2}{*}{MXFP6} & FP6 (E3M2) & \multirow{2}{*}{6} &  &  \\
 & FP6 (E2M3) &  &  &  \\ \cline{1-3}
MXFP4 & FP4 (E2M1) & 4 &  &  \\ \cline{1-3}
MXINT8 & INT8 & 8 &  &  \\ \hline
\end{tabular}%
}
\caption{MX-compliant format. Configurations are adapted from~\cite{ocp}.}
\label{tab:mx_configs}
\end{table}

%% file: Sections/appendix-experiments.tex
\section{Experimental Details}
\label{appendix:experimental_deatils}
\textbf{Quantization Settings.} Our experiments is conducted by modifying the PyTorch and CUDA code within the \texttt{MX Emulation library}~\cite{rouhani2023microscaling}. We quantize all weights and activations in Transformer decoder layers, including \textbf{\textit{Query}, \textit{Key}, \textit{Self-attention map}, and \textit{Value}} as a default.

\textbf{Models.}
The models used in the experiments include OPT~\cite{opt}, LLaMA~\cite{touvron2023llama2},~\cite{llama3modelcard}, Qwen~\cite{bai2023qwen}, and Mistral~\cite{jiang2023mistral}, LLaMA2-Chat~\cite{touvron2023llama2}, BART~\cite{lewis2019bart}, and LLaVA~\cite{liu2023llava} (which backbone is Vicuna-7B~\cite{vicuna2023}).

\textbf{Robustness Measurment Settings in Table~\ref{tab:overfitting}.}
Following the calibration robustness measurement method introduced in AWQ~\cite{lin2023awq}, we select two subsets from the Pile dataset~\cite{gao2020pile}: PubMed Abstracts~\cite{pubmeddataset} and Enron Emails~\cite{enron}. The calibration and evaluation sets are distinct, with no overlap; 128 samples with a sequence length of 2048 are used for calibration, and 200 samples are reserved for perplexity evaluation. Additionally, we configure the calibration set with questions and answers from the PIQA~\cite{bisk2019piqa} and WinoGrande~\cite{sakaguchi2019winogrande} datasets to analyze calibration effects in question-answering tasks. To determine whether our improved MX format can effectively replace existing techniques for W4A4 inference, we align the experimental settings, applying reduced-precision activations consistent with prior studies (excluding quantization for Query and Softmax output). We reproduce the performance of QuaRot and SpinQuant following their official repositories, with modifications to calibration and evaluation datasets.

\textbf{MT-Bench.}  MT-Bench assigns scores ranging from 1 to 10, given by GPT-4~\cite{gpt4}, to responses generated from an initial question and a subsequent follow-up question across 80 multi-turn conversations. 

\textbf{Visual Tasks.}
For evaluating VLMs, we utilize \texttt{lmms-eval}~\cite{zhang2024lmmsevalrealitycheckevaluation}, including TextVQA (VQA-T)~\cite{singh2019towardstextvqa}, DocVQA~\cite{mathew2021docvqadatasetvqadocument}, OCRBench~\cite{liu2024ocrbenchhiddenmysteryocr}, and ChartQA~\cite{masry-etal-2022-chartqa}.

\textbf{Long-Context Benchmarks.}
To measure the effectiveness of AMXFP4 while long-context is given, we utilize LongBench-E~\cite{bai-etal-2024-longbench} on LLaMA2-Chat-7B. LongBench-E includes 13 tasks: Qasper~\cite{qasper}, MultiFieldQA~\cite{bai-etal-2024-longbench}, HotPotQA~\cite{yang2018hotpotqadatasetdiverseexplainable}, MultihopQA~\cite{ho-etal-2020-constructing-multihop}, GovReport~\cite{huang-etal-2021-efficient-govreport}, MultiNews~\cite{bai-etal-2024-longbench}, TREC~\cite{li-roth-2002-learning-trec}, TriviaQA~\cite{2017arXivtriviaqa}, SAMSum~\cite{gliwa-etal-2019-samsum}, PassageCount~\cite{bai-etal-2024-longbench}, PassageRetrieval~\cite{bai-etal-2024-longbench}, LCC~\cite{guo2023longcoderlongrangepretrainedlanguage_lcc}, and RepoBench-P~\cite{liu2023repobenchbenchmarkingrepositorylevelcode}.

\textbf{Knowledge Evaluation Benchmarks (MMLU and CSQA).}
We evaluate our method into commonsence QA (CSQA) (PIQA~\cite{Bisk2020piqa}, WinoGrande~\cite{sakaguchi2019winogrande}, ARC challenge~\cite{allenai:arc}) and MMLU~\cite{mmlu}. and CSQA and MMLU accuracies are assessed using the \texttt{lm-evaluation-harness}~\cite{eval-harness}.

\textbf{Quantization-Aware Training Settings.}
We conduct quantization-aware training (QAT) experiments on LLaMA3-8B, specifically because it exhibits relatively high perplexity degradation under direct-cast quantization. We quantize all linear layer weights and activations to 4 bits and employ flash-attention~\cite{dao2022flashattention} for attention operations. We construct QAT dataset by randomly sampling 3200 sequences, each with a length of 2048 tokens (a total of 6.5M tokens), from the Wikitext-2 training set. Training is performed for 100 steps with an effective batch size of 32, and we search learning rates between 2e-6 and 1e-5 to determine the best hyperparameters for both MXFP4 and AMXFP4.

\section{More Experimental Results}
\label{appendix:experimental_results}

\begin{figure}[t]
\centering
\centerline{\includegraphics[width=\columnwidth]{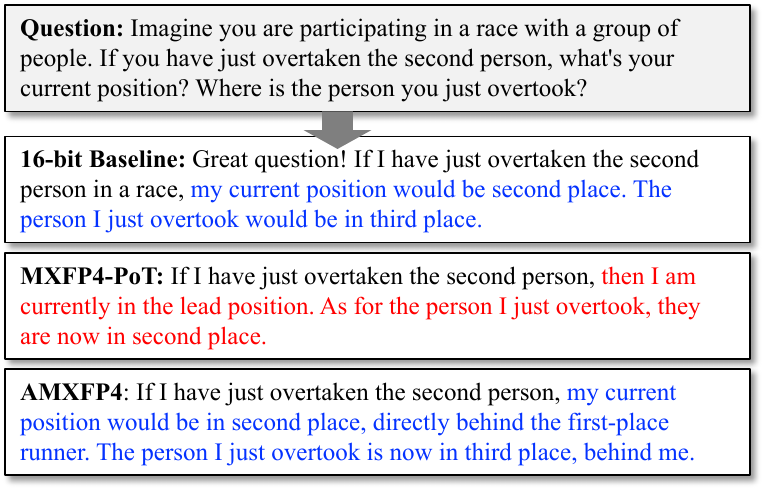}}
\caption{Example of chatbot interactions from MT-Bench (LLaMA2-Chat-7B)}
\label{fig:chat-example}
\end{figure}

\subsection{Ablation Studies}
\label{appendix:ablation}

\input{Tables/mx-ppl}
\textbf{Language Modeling Tasks.} We evaluate on language modeling with WikiText~\cite{merity2016pointerwiki}. The perplexity measurement on the Wikitext test dataset involves grouping 2048 tokens collectively. Table~\ref{tab:mx-ppl} presents Wikitext-2 perplexity results for six LLMs across MXFP4 and AMXFP4 with PoT and FP8 shared scale. While MXFP4-PoT introduces significant perplexity degradation across all models, employing MXFP4 with an enhanced shared scale substantially reduces perplexity in each case. Notably, AMXFP4, through asymmetric data representation, achieves a 0.59 perplexity reduction in LLaMA3-8B compared to MXFP4 and limits perplexity degradation to only about 0.46 in models like Mistral-7B.

\input{Tables/mx-summarization}
\textbf{Encoder-Decoder Language Model.} Table~\ref{tab:mx-summarization} displays the ROUGE~\cite{lin-2004-rouge} scores for BART-Large’s~\cite{lewis2019bart} summarization task on the CNN/DailyMail dataset~\cite{see-etal-2017-get-cnn} across different MX format options. AMXFP4 exhibits only a 0.7-point drop in ROUGE-L score compared to the baseline, demonstrating that the proposed data format also enables effective 4-bit inference in encoder-decoder models.

\begin{table}[t]
\centering
    \resizebox{0.65\columnwidth}{!}{%
    \begin{tabular}{cc}
    \hline
    {MX Format} & {Wikitext-2 Perplexity $\downarrow$} \\
    \hline
    MXINT4   & 7.73 \\
    AMXINT4  & 5.36 \\
    MXFP4    & 5.82 \\
    \rowcolor[HTML]{ECF4FF} 
    AMXFP4   & \textbf{4.35} \\
    \hline
    \end{tabular}
    }
\caption{LLaMA3-70B perplexity on Wikitext-2 inference across various MX configurations. FP16 baseline perplexity is 2.86.}
\label{tab:llama3-wikitext}
\end{table}

\textbf{LLaMA3-70B Evaluation on Wikitext-2}
To validate the scalability and practical utility of our proposed AMX formats on larger-scale models, we additionally evaluate AMXINT4 and AMXFP4 on the LLaMA3-70B model using Wikitext-2 inference. As shown in Table~\ref{tab:llama3-wikitext}, AMXFP4 continues to demonstrate superior performance compared to prior MX configurations. MXINT4 exhibits the highest perplexity (7.73), while MXFP4 reduces it to 5.82. AMXINT4 further improves performance, achieving 5.36, consistent with the element-wise format selection method described in Section~\ref{sec4.1}. AMXFP4 achieves the lowest perplexity (4.35), significantly outperforming all other formats. These results underscore the effectiveness of addressing data asymmetry in microscaling, particularly in the context of large-scale models such as LLaMA3-70B.

\begin{table*}[t]
    \centering
    \resizebox{0.8\textwidth}{!}{%
    \begin{tabular}{l l c c c c c c}
        \toprule
        \textbf{Pruning Ratio} & \textbf{Bit-Configurations} & \textbf{Memory (GB)} & \textbf{BoolQ} & \textbf{OBQA} & \textbf{PIQA} & \textbf{ARC-C} & \textbf{Average} $\uparrow$ \\
        \midrule
        0\%  & FP16 & 13.48 & 75.11 & 44.40 & 79.16 & 44.71 & 60.85 \\
        \midrule
        20\% & FP16 & 10.85 & 66.45 & 41.40 & 78.13 & 39.42 & 56.35 \\
        % 20\% & MXFP4 & 3.27 & 58.07 & 36.80 & 72.85 & 34.39 & 50.53 \\
        % 20\% & Ours-FP4Asym-bs32-PoT & 3.27 & 57.43 & 39.00 & 72.85 & 34.13 & 50.85 \\
        20\% & MXFP4-PoT & 3.27 & 61.74 & 36.80 & 73.39 & 35.15 & 51.77 \\
        % 20\% & Ours-FP4Asym-bs32-PoTR & 3.27 & 60.83 & 38.00 & 73.72 & 35.49 & 52.01 \\
        20\% & MXFP4 & 3.27 & 62.91 & 37.60 & 75.19 & 36.77 & 53.12 \\
        20\% & AMXFP4 & 3.27 & 62.72 & 38.60 & 75.73 & 36.43 & 53.37 \\
        \bottomrule
    \end{tabular}
    }
    \caption{Performance comparison across different pruning ratios and bit configurations (LLaMA-7B).}
    \label{tab:pruning_comparison}
\end{table*}

\textbf{Conjunction with Sparsity}
We conduct an ablation study by applying MXFP4 to a pruned model to see if improvements in the micro-scaled reduced-precision option can work in conjunction with other methods like sparsity. We use 20\% pruning with LLM-Pruner~\cite{ma2023llmpruner} as the baseline for the sparse model.
Table~\ref{tab:pruning_comparison} shows the accuracy when applying various MXFP4 options to the pruned model for four CSQA tasks. The model with 20\% pruning reduces the requried memory while tolerating a slight drop in accuracy. Applying MXFP4-PoT to the pruned model results in an additional 5\% performance drop. On the other hand, advancements in shared scale and the representation of asymmetric data have progressively enhanced accuracy even in pruned models, showing that the improvements of the proposed MX format have a cumulative effect.

\textbf{Ablation Study on Shared-Scale Bit-Encoding.} Table~\ref{tab:shared_scale_ablation} illustrates the perplexity according to the type of shared scale across various models and group sizes. In the case of FP4, using the default 8-bit PoT (Floor) shared scale option of MX, there is a notable increase in perplexity as the group size decreases. This trend is also observed in AsymFP4, primarily due to the increased error from frequent clamping caused by the Floor operation. To address this, our proposed 8-bit PoT consistently improves performance even with smaller group sizes. On the other hand, FP8, another 8-bit alternative, with a 4-bit exponent, significantly degrades performance in models like Mistral, a consequence of its inherent limitations in dynamic range. Conversely, our findings demonstrate that using a 5-bit exponent FP8 shared scale can achieve performance close to FP16.

\subsection{Detailed Results for Practical Applications}

\begin{figure*}[htbp]
\centering
\centerline{\includegraphics[width=\textwidth]{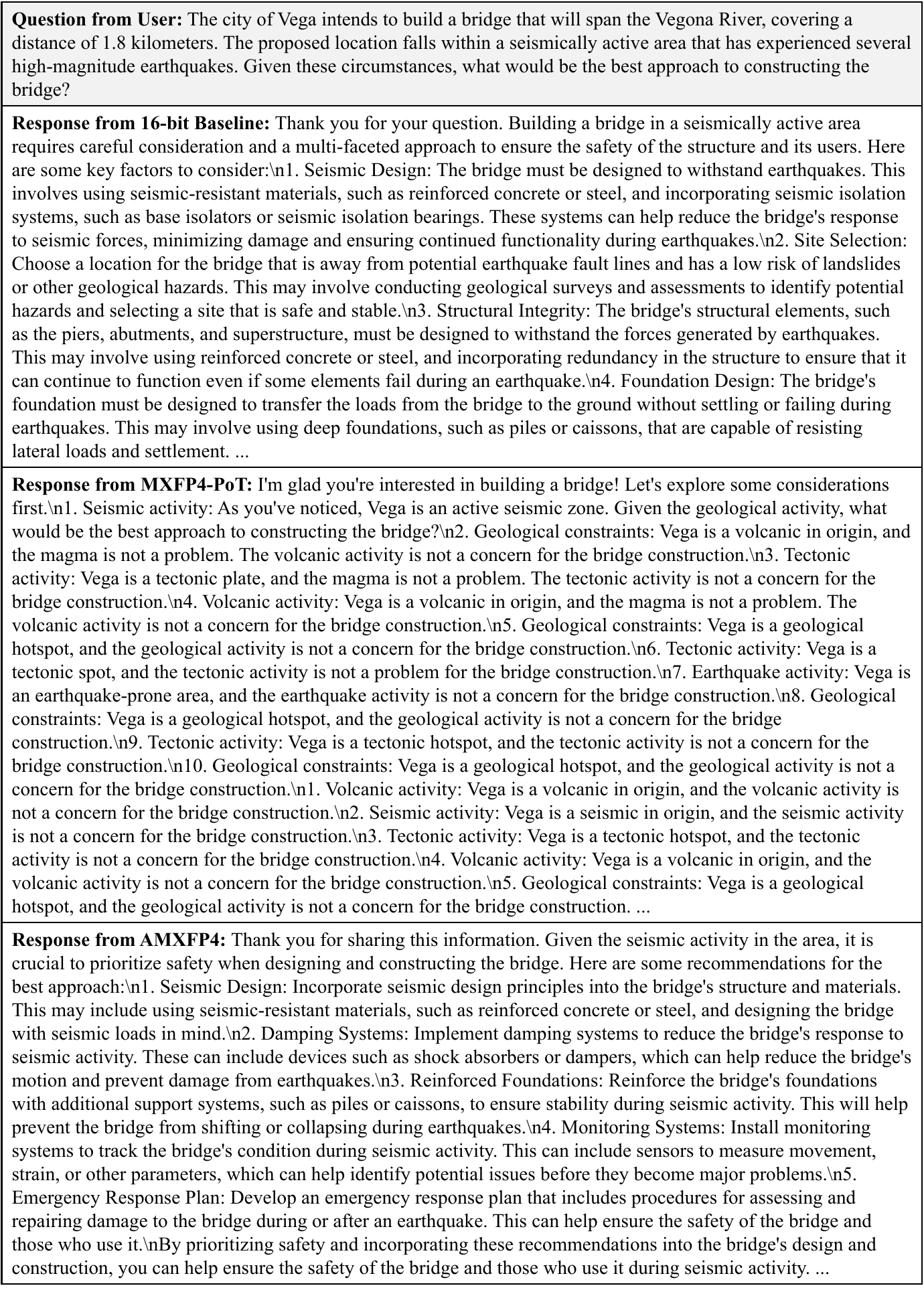}}
\caption{MT-Bench example (LLaMA2-Chat-7B).}
\label{fig:mt-bench-example}
\end{figure*}

\input{Tables/mx-mtbench}
\textbf{Chatbot Results.}
Fig.~\ref{fig:mt-bench-example} presents an example from MT-Bench. While the 16-bit baseline provides responses aligned with the user’s intent, MXFP4 tends to generate repetitive and unhelpful sentences. In contrast, AMXFP4 produces responses that, similar to the baseline, are useful to the user. Table~\ref{tab:mx-mtbench} displays the single scores from MT-Bench across different categories. The proposed AMXFP4 demonstrates the ability to recover baseline performance in most sub-categories.

\begin{figure}[t]
\centering
\centerline{\includegraphics[width=\columnwidth]{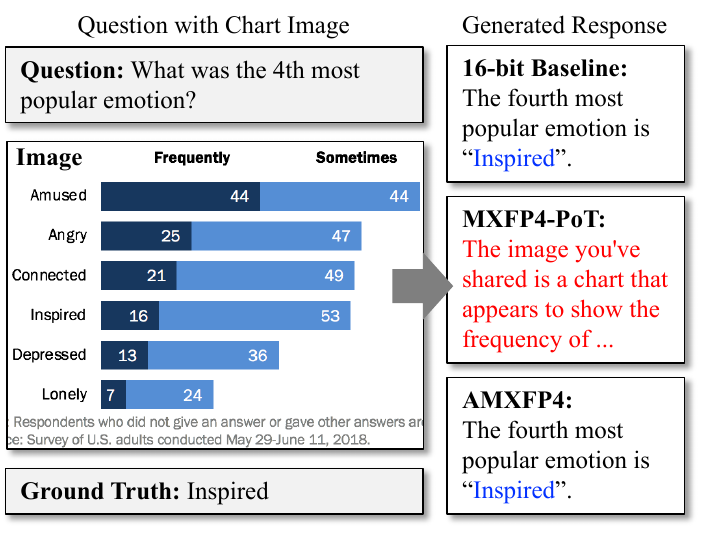}}
\caption{Comparison between responses from MXFP4-PoT and AMXFP4 in ChartQA example.}
\label{fig:chartqa}
\end{figure}
\textbf{Visual Question Answering Results.}
Fig.~\ref{fig:chartqa} presents an example response to a given chart image using MXFP4 and AMXFP4. While MXFP4-PoT generates irrelevant responses, AMXFP4 produces the correct ground-truth answer, identical to the baseline.

\input{Tables/mx-longbench}
\textbf{LongBench-E Results.}
Table~\ref{tab:longbench} provides detailed scores for 13 benchmarks in LongBench-E. MXFP4-PoT shows significant performance degradation overall, with a score drop of 16 for contexts exceeding 8K. While MXFP4 substantially improves over MXFP4-PoT, it still experiences a score reduction of over 6 when handling contexts above 8K. AMXFP4 increases the average score by more than 2 compared to MXFP4 for long-context scenarios exceeding 8K.

\input{Tables/shared_scale_ablation}
\input{Tables/cluster-mse}

%% file: Tables/mx-ppl.tex
\begin{table}[t]
\centering
\resizebox{\columnwidth}{!}{%
\begin{tabular}{c|cc|ccc|c}
\hline
\multirow{2}{*}{Data Format} & \multicolumn{2}{c|}{OPT} & \multicolumn{3}{c|}{LLaMA} & Mistral \\ \cline{2-7} 
 & 6.7B & 13B & 2-7B & 2-13B & 3-8B & 7B \\ \hline
16-bit Baseline & 10.86 & 10.13 & 5.47 & 4.88 & 6.14 & 5.25 \\ \hline
MXFP4-PoT & 25.51 & 12.88 & 7.83 & 6.98 & 11.17 & 6.34 \\
MXFP4 & 13.71 & 12.09 & 6.49 & 5.69 & 8.31 & 5.88 \\ \hline
% AMXFP4-PoT & 25.00 & 12.65 & 7.46 & 6.60 & 10.05 & 6.19 \\
\rowcolor[HTML]{ECF4FF} 
AMXFP4 & \textbf{13.06} & \textbf{11.90} & \textbf{6.22} & \textbf{5.47} & \textbf{7.72} & \textbf{5.71} \\ \hline
\end{tabular}%
}
\caption{Wikitext-2 inference for MXFP4 and AMXFP4.}
\label{tab:mx-ppl}
\end{table}

%% file: Tables/mx-summarization.tex
% \usepackage{graphicx}
\begin{table}[t]
\centering
\resizebox{\columnwidth}{!}{%
\begin{tabular}{c|ccc}
\hline
Data Format & ROUGE-1 ↑ & ROUGE-2 ↑ & ROUGE-L ↑ \\ \hline
16-bit Baseline & 45.09 & 21.60 & 31.43 \\ \hline
MXFP4-PoT & 42.47 & 19.10 & 29.18 \\
MXFP4 & 43.73 & 20.50 & 30.43 \\
% AMXFP4-PoT & 43.06 & 19.59 & 29.66 \\
\rowcolor[HTML]{ECF4FF} 
AMXFP4 & \textbf{44.13} & \textbf{20.79} & \textbf{30.72} \\ \hline
\end{tabular}%
}
\caption{CNN/DailyMail summarization task on BART-Large.}
\label{tab:mx-summarization}
\end{table}

%% file: Tables/mx-mtbench.tex
% \usepackage{graphicx}
% \usepackage[normalem]{ulem}
\begin{table*}[t]
\centering
\resizebox{0.9\textwidth}{!}{%
\begin{tabular}{c|cccccccc|c}
\hline
{\color[HTML]{000000} Data Format} & {\color[HTML]{000000} Writing} & {\color[HTML]{000000} Roleplay} & {\color[HTML]{000000} Reasoning} & {\color[HTML]{000000} Math} & {\color[HTML]{000000} Coding} & {\color[HTML]{000000} Extraction} & {\color[HTML]{000000} STEM} & {\color[HTML]{000000} Humanities} & {\color[HTML]{000000} Single Score} \\ \hline
{\color[HTML]{000000} 16-bit Baseline} & {\color[HTML]{000000} 9.25} & {\color[HTML]{000000} 7.20} & {\color[HTML]{000000} 4.65} & {\color[HTML]{000000} 2.55} & {\color[HTML]{000000} 3.30} & {\color[HTML]{000000} 5.55} & {\color[HTML]{000000} 8.93} & {\color[HTML]{000000} 9.58} & {\color[HTML]{000000} 6.38} \\ \hline
{\color[HTML]{000000} MXFP4-PoT} & {\color[HTML]{000000} 4.30} & {\color[HTML]{000000} 4.05} & {\color[HTML]{000000} 2.35} & {\color[HTML]{000000} 1.90} & {\color[HTML]{000000} 1.25} & {\color[HTML]{000000} 1.55} & {\color[HTML]{000000} 5.23} & {\color[HTML]{000000} 5.15} & {\color[HTML]{000000} 3.22} \\
{\color[HTML]{000000} MXFP4} & {\color[HTML]{000000} 7.20} & {\color[HTML]{000000} \textbf{7.03}} & {\color[HTML]{000000} 3.95} & {\color[HTML]{000000} 1.70} & {\color[HTML]{000000} 1.70} & {\color[HTML]{000000} 4.35} & {\color[HTML]{000000} 7.53} & {\color[HTML]{000000} 8.53} & {\color[HTML]{000000} {\ul 5.25}} \\ \hline
\rowcolor[HTML]{ECF4FF} 
{\color[HTML]{000000} AMXFP4} & {\color[HTML]{000000} \textbf{8.20}} & {\color[HTML]{000000} 5.98} & {\color[HTML]{000000} \textbf{4.50}} & {\color[HTML]{000000} \textbf{2.50}} & {\color[HTML]{000000} \textbf{3.05}} & {\color[HTML]{000000} \textbf{5.16}} & {\color[HTML]{000000} \textbf{7.70}} & {\color[HTML]{000000} \textbf{8.70}} & {\color[HTML]{000000} \textbf{5.73}} \\ \hline
\end{tabular}%
}
\caption{MT-Bench Single Score (LLaMA2-Chat-7B).}
\label{tab:mx-mtbench}
\end{table*}

%% file: Tables/mx-longbench.tex
% \usepackage{multirow}
% \usepackage{graphicx}
\begin{table*}[t]
\centering
\resizebox{\textwidth}{!}{%
\begin{tabular}{c|c|cc|cc|cc|ccc|cc|cc|c}
\hline
 &  & \multicolumn{2}{c|}{Single Doc-QA} & \multicolumn{2}{c|}{Multi Doc-QA} & \multicolumn{2}{c|}{Summarization} & \multicolumn{3}{c|}{Few-shot Learning} & \multicolumn{2}{c|}{Synthetic Tasks} & \multicolumn{2}{c|}{Code Completion} &  \\ \cline{3-15}
\multirow{-2}{*}{Data Format} & \multirow{-2}{*}{\begin{tabular}[c]{@{}c@{}}Context\\ Length\end{tabular}} & Qasper & \begin{tabular}[c]{@{}c@{}}MultiField\\ QA\end{tabular} & \begin{tabular}[c]{@{}c@{}}Hotpot\\ QA\end{tabular} & \begin{tabular}[c]{@{}c@{}}Multihop\\ QA\end{tabular} & \begin{tabular}[c]{@{}c@{}}Gov\\ Report\end{tabular} & \begin{tabular}[c]{@{}c@{}}Multi\\ News\end{tabular} & TREC & \begin{tabular}[c]{@{}c@{}}Trivia\\ QA\end{tabular} & \begin{tabular}[c]{@{}c@{}}SAM\\ Sum\end{tabular} & \begin{tabular}[c]{@{}c@{}}Passage\\ Count\end{tabular} & \begin{tabular}[c]{@{}c@{}}Passage\\ Retrieval\end{tabular} & LCC & \begin{tabular}[c]{@{}c@{}}Repo\\ Bench-P\end{tabular} & \multirow{-2}{*}{Average} \\ \hline
 & 0-4k & 22.99 & 43.37 & 37.14 & 35.79 & 31.13 & 26.84 & 54.00 & 83.13 & 39.33 & 6.35 & 18.00 & 62.45 & 49.02 & 39.20 \\
 & 4-8k & 18.37 & 32.29 & 30.47 & 24.36 & 27.89 & 23.14 & 60.00 & 84.02 & 37.73 & 2.01 & 4.00 & 59.98 & 48.05 & 34.79 \\
\multirow{-3}{*}{16-bit Baseline} & 8k+ & 21.42 & 25.59 & 24.08 & 23.37 & 25.14 & 23.11 & 60.00 & 91.51 & 40.22 & 2.72 & 7.00 & 56.88 & 48.51 & 34.58 \\ \hline
 & 0-4k & 12.02 & 31.91 & 14.27 & 15.82 & 20.23 & 20.16 & 32.00 & 44.39 & 28.37 & 4.48 & 9.42 & 31.54 & 34.96 & 23.04 \\
 & 4-8k & 11.02 & 17.56 & 13.83 & 13.32 & 15.71 & 13.96 & 37.00 & 36.66 & 25.93 & 6.07 & 2.12 & 32.13 & 32.50 & 19.83 \\
\multirow{-3}{*}{MXFP4-PoT} & 8k+ & 9.27 & 10.26 & 10.78 & 10.10 & 13.94 & 13.13 & 36.00 & 41.83 & 24.92 & 5.72 & 5.09 & 27.31 & 35.29 & 18.74 \\
 & 0-4k & 13.16 & 40.81 & 25.27 & 24.27 & 22.68 & 23.66 & 46.00 & 77.49 & 38.97 & 5.71 & 9.98 & 49.54 & 41.24 & 32.21 \\
 & 4-8k & 14.26 & 27.40 & 21.96 & 19.36 & 19.91 & 18.59 & 58.00 & 75.53 & 35.98 & 1.50 & 0.79 & 48.15 & 38.45 & 29.22 \\
\multirow{-3}{*}{MXFP4} & 8k+ & 10.04 & 23.07 & 19.15 & 17.19 & 18.09 & 18.66 & 49.00 & 79.39 & 37.82 & 3.68 & 5.00 & 45.10 & 41.77 & 28.30 \\
\rowcolor[HTML]{ECF4FF} 
\cellcolor[HTML]{ECF4FF}{\color[HTML]{000000} } & {\color[HTML]{000000} 0-4k} & {\color[HTML]{000000} 16.93} & {\color[HTML]{000000} 34.62} & {\color[HTML]{000000} 32.16} & {\color[HTML]{000000} 25.52} & {\color[HTML]{000000} 23.21} & {\color[HTML]{000000} 23.49} & {\color[HTML]{000000} 50.00} & {\color[HTML]{000000} 76.52} & {\color[HTML]{000000} 37.88} & {\color[HTML]{000000} 9.81} & {\color[HTML]{000000} 10.50} & {\color[HTML]{000000} 50.76} & {\color[HTML]{000000} 43.95} & {\color[HTML]{000000} \textbf{33.49}} \\
\rowcolor[HTML]{ECF4FF} 
\cellcolor[HTML]{ECF4FF}{\color[HTML]{000000} } & {\color[HTML]{000000} 4-8k} & {\color[HTML]{000000} 19.56} & {\color[HTML]{000000} 26.96} & {\color[HTML]{000000} 26.03} & {\color[HTML]{000000} 19.74} & {\color[HTML]{000000} 19.80} & {\color[HTML]{000000} 19.71} & {\color[HTML]{000000} 54.00} & {\color[HTML]{000000} 70.53} & {\color[HTML]{000000} 36.29} & {\color[HTML]{000000} 2.04} & {\color[HTML]{000000} 5.27} & {\color[HTML]{000000} 48.05} & {\color[HTML]{000000} 40.77} & {\color[HTML]{000000} \textbf{29.90}} \\
\rowcolor[HTML]{ECF4FF} 
\multirow{-3}{*}{\cellcolor[HTML]{ECF4FF}{\color[HTML]{000000} AMXFP4}} & {\color[HTML]{000000} 8k+} & {\color[HTML]{000000} 34.32} & {\color[HTML]{000000} 17.40} & {\color[HTML]{000000} 20.52} & {\color[HTML]{000000} 21.72} & {\color[HTML]{000000} 18.04} & {\color[HTML]{000000} 18.70} & {\color[HTML]{000000} 50.00} & {\color[HTML]{000000} 79.92} & {\color[HTML]{000000} 38.73} & {\color[HTML]{000000} 3.39} & {\color[HTML]{000000} 9.00} & {\color[HTML]{000000} 45.12} & {\color[HTML]{000000} 40.50} & {\color[HTML]{000000} \textbf{30.57}} \\ \hline
\end{tabular}%
}
\caption{Detailed scores of LongBench-E~\cite{bai-etal-2024-longbench}.}
\label{tab:longbench}
\end{table*}

%% file: Tables/shared_scale_ablation.tex
% \usepackage{multirow}
% \usepackage{graphicx}
\begin{table*}[t]
\centering
\resizebox{0.8\textwidth}{!}{%
\begin{tabular}{ccc|cc|cc|c|c|c}
\hline
\multicolumn{1}{c|}{\multirow{2}{*}{\begin{tabular}[c]{@{}c@{}}Data\\ Format\end{tabular}}} & \multicolumn{1}{c|}{\multirow{2}{*}{\begin{tabular}[c]{@{}c@{}}Shared\\ Scale\end{tabular}}} & \multirow{2}{*}{\begin{tabular}[c]{@{}c@{}}Group\\ Size\end{tabular}} & \multicolumn{2}{c|}{OPT} & \multicolumn{2}{c|}{LLaMA2} & LLaMA3 & Mistral 7B & Qwen \\
\multicolumn{1}{c|}{} & \multicolumn{1}{c|}{} &  & 6.7B & 13B & 7B & 13B & 8B & 7B & 7B \\ \hline
\multicolumn{3}{c|}{16-bit Baseline} & 10.860 & 10.128 & 5.472 & 4.884 & 6.137 & 5.252 & 7.605 \\ \hline
\multicolumn{1}{c|}{\multirow{20}{*}{MXFP4}} & \multicolumn{1}{c|}{\multirow{4}{*}{FP16}} & 128 & 12.566 & 12.415 & 7.065 & 6.208 & 9.826 & 6.137 & 8.669 \\
\multicolumn{1}{c|}{} & \multicolumn{1}{c|}{} & 64 & 11.843 & 11.958 & 6.470 & 5.667 & 8.368 & 5.854 & 8.364 \\
\multicolumn{1}{c|}{} & \multicolumn{1}{c|}{} & 32 & 11.475 & 11.084 & 6.206 & 5.444 & 7.851 & 5.722 & 8.214 \\
\multicolumn{1}{c|}{} & \multicolumn{1}{c|}{} & 16 & 11.233 & 10.841 & 6.015 & 5.284 & 7.334 & 5.607 & 8.084 \\ \cline{2-10} 
\multicolumn{1}{c|}{} & \multicolumn{1}{c|}{\multirow{4}{*}{PoT (Floor)}} & 128 & 24.126 & 16.151 & 12.056 & 11.243 & 17.848 & 8.454 & 10.407 \\
\multicolumn{1}{c|}{} & \multicolumn{1}{c|}{} & 64 & 22.605 & 14.820 & 11.228 & 10.453 & 16.636 & 8.846 & 10.023 \\
\multicolumn{1}{c|}{} & \multicolumn{1}{c|}{} & 32 & 22.525 & 14.473 & 11.150 & 10.270 & 16.636 & 9.454 & 9.762 \\
\multicolumn{1}{c|}{} & \multicolumn{1}{c|}{} & 16 & 23.463 & 14.638 & 11.212 & 10.065 & 18.582 & 10.392 & 9.651 \\ \cline{2-10} 
\multicolumn{1}{c|}{} & \multicolumn{1}{c|}{\multirow{4}{*}{PoT (Round)}} & 128 & 40.288 & 14.460 & 9.383 & 8.472 & 15.741 & 7.000 & 9.635 \\
\multicolumn{1}{c|}{} & \multicolumn{1}{c|}{} & 64 & 27.696&13.238&8.393&7.669&12.450&6.585&9.185 \\
\multicolumn{1}{c|}{} & \multicolumn{1}{c|}{} & 32 & 25.512&12.879&7.834&6.982&11.171&6.337&8.940 \\
\multicolumn{1}{c|}{} & \multicolumn{1}{c|}{} & 16 & 25.155&12.683&7.495&6.649&10.381&6.206&8.764 \\ \cline{2-10} 
\multicolumn{1}{c|}{} & \multicolumn{1}{c|}{\multirow{4}{*}{FP8 (1-4-3)}} & 128 & 21.914 & 14.075 & 10.749 & 9.883 & 9.842 & 55.719 & 8.783 \\
\multicolumn{1}{c|}{} & \multicolumn{1}{c|}{} & 64 & 18.637 & 15.840 & 11.036 & 9.340 & 8.761 & 670.647 & 8.458 \\
\multicolumn{1}{c|}{} & \multicolumn{1}{c|}{} & 32 & 24.109 & 21.447 & 13.334 & 9.705 & 8.733 & 6050.050 & 8.358 \\
\multicolumn{1}{c|}{} & \multicolumn{1}{c|}{} & 16 & 28.186 & 33.131 & 17.082 & 11.330 & 8.340 & 25756.484 & 8.229 \\ \cline{2-10} 
\multicolumn{1}{c|}{} & \multicolumn{1}{c|}{\multirow{4}{*}{FP8 (1-5-2)}} & 128 & 15.857 & 14.530 & 7.390 & 6.450 & 10.408 & 6.234 & 8.806 \\
\multicolumn{1}{c|}{} & \multicolumn{1}{c|}{} & 64 & 14.075 & 12.777 & 6.788 & 5.923 & 8.952 & 5.957 & 8.542 \\
\multicolumn{1}{c|}{} & \multicolumn{1}{c|}{} & 32 & 13.712 & 12.091 & 6.490 & 5.691 & 8.307 & 5.883 & 8.366 \\
\multicolumn{1}{c|}{} & \multicolumn{1}{c|}{} & 16 & 13.534 & 11.808 & 6.265 & 5.520 & 7.824 & 5.725 & 8.247 \\ \hline
\multicolumn{1}{c|}{\multirow{20}{*}{AMXFP4}} & \multicolumn{1}{c|}{\multirow{4}{*}{FP16}} & 128 & 12.107 & 11.718 & 6.564 & 5.712 & 8.364 & 5.898 & 8.408 \\
\multicolumn{1}{c|}{} & \multicolumn{1}{c|}{} & 64 & 11.489 & 11.187 & 6.173 & 5.400 & 7.660 & 5.702 & 8.272 \\
\multicolumn{1}{c|}{} & \multicolumn{1}{c|}{} & 32 & 11.242 & 10.900 & 5.999 & 5.261 & 7.296 & 5.588 & 8.066 \\
\multicolumn{1}{c|}{} & \multicolumn{1}{c|}{} & 16 & 11.118 & 10.581 & 5.840 & 5.149 & 6.978 & 5.507 & 7.953 \\ \cline{2-10} 
\multicolumn{1}{c|}{} & \multicolumn{1}{c|}{\multirow{4}{*}{PoT (Floor)}} & 128 & 23.161 & 15.074 & 11.555 & 10.839 & 18.404 & 8.594 & 10.123 \\
\multicolumn{1}{c|}{} & \multicolumn{1}{c|}{} & 64 & 24.002 & 14.635 & 10.956 & 10.380 & 18.910 & 9.217 & 9.840 \\
\multicolumn{1}{c|}{} & \multicolumn{1}{c|}{} & 32 & 25.233 & 14.569 & 11.362 & 10.433 & 18.748 & 10.710 & 9.584 \\
\multicolumn{1}{c|}{} & \multicolumn{1}{c|}{} & 16 & 27.992 & 14.910 & 12.255 & 11.118 & 22.084 & 14.090 & 9.595 \\ \cline{2-10} 
\multicolumn{1}{c|}{} & \multicolumn{1}{c|}{\multirow{4}{*}{PoT (Round)}} & 128 & 28.781 & 13.485 & 8.454 & 7.466 & 12.307 & 6.517 & 9.235 \\
\multicolumn{1}{c|}{} & \multicolumn{1}{c|}{} & 64 & 26.021&12.939&7.803&7.002&10.683&6.311&8.987 \\
\multicolumn{1}{c|}{} & \multicolumn{1}{c|}{} & 32 & 24.995&12.651&7.456&6.596&10.048&6.189&8.780 \\
\multicolumn{1}{c|}{} & \multicolumn{1}{c|}{} & 16 & 24.240&12.585&7.172&6.362&9.688&6.120&8.673 \\ \cline{2-10} 
\multicolumn{1}{c|}{} & \multicolumn{1}{c|}{\multirow{4}{*}{FP8 (1-4-3)}} & 128 & 17.243 & 13.764 & 9.725 & 8.966 & 8.640 & 1053.763 & 8.468 \\
\multicolumn{1}{c|}{} & \multicolumn{1}{c|}{} & 64 & 18.093 & 16.331 & 10.582 & 8.622 & 8.609 & 3718.406 & 8.303 \\
\multicolumn{1}{c|}{} & \multicolumn{1}{c|}{} & 32 & 20.803 & 22.674 & 13.080 & 9.435 & 8.193 & 13421.343 & 8.231 \\
\multicolumn{1}{c|}{} & \multicolumn{1}{c|}{} & 16 & 31.017 & 40.884 & 17.459 & 11.331 & 8.260 & 30513.367 & 8.175 \\ \cline{2-10} 
\multicolumn{1}{c|}{} & \multicolumn{1}{c|}{\multirow{4}{*}{FP8 (1-5-2)}} & 128 & 14.580 & 12.652 & 6.847 & 5.901 & 8.777 & 6.003 & 8.568 \\
\multicolumn{1}{c|}{} & \multicolumn{1}{c|}{} & 64 & 13.480 & 12.132 & 6.451 & 5.618 & 8.092 & 5.817 & 8.400 \\
\multicolumn{1}{c|}{} & \multicolumn{1}{c|}{} & 32 & 13.058 & 11.902 & 6.223 & 5.469 & 7.725 & 5.707 & 8.215 \\
\multicolumn{1}{c|}{} & \multicolumn{1}{c|}{} & 16 & 12.941 & 11.625 & 6.064 & 5.374 & 7.421 & 5.632 & 8.114 \\ \hline
\end{tabular}%
}
\caption{Ablation study on shared scale bit-encoding.}
\label{tab:shared_scale_ablation}
\end{table*}

%% file: Tables/cluster-mse.tex
\begin{table*}[t]
\centering
\resizebox{\textwidth}{!}{%
\begin{tabular}{ccc|ccccccc}
\hline
\multicolumn{1}{c|}{\multirow{2}{*}{Cluster ID}} & \multicolumn{2}{c|}{Centroids} & \multicolumn{7}{c}{Data Formats} \\ \cline{2-10} 
\multicolumn{1}{c|}{} & Normalized Mean & Normalized Kurtosis & NF4 & SF4 & INT4 & Asym INT4 & FP4 & \multicolumn{1}{c|}{Asym FP4} & Lloyd-Max \\ \hline
\multicolumn{1}{c|}{0} & 0.041 & 0.003 & 4.14E-04 & 5.24E-04 & 5.77E-04 & 3.90E-04 & 5.45E-04 & \multicolumn{1}{c|}{4.65E-04} & 3.85E-04 \\
\multicolumn{1}{c|}{1} & -0.084 & 0.472 & 2.63E-03 & 1.86E-03 & 7.06E-03 & 2.41E-03 & 2.42E-03 & \multicolumn{1}{c|}{1.43E-03} & 8.07E-04 \\
\multicolumn{1}{c|}{2} & -0.357 & -0.010 & 4.18E-04 & 5.70E-04 & 4.80E-04 & 3.17E-04 & 5.40E-04 & \multicolumn{1}{c|}{4.77E-04} & 3.30E-04 \\
\multicolumn{1}{c|}{3} & 0.533 & -0.016 & 3.72E-04 & 5.27E-04 & 4.16E-04 & 2.68E-04 & 5.44E-04 & \multicolumn{1}{c|}{4.91E-04} & 2.89E-04 \\
\multicolumn{1}{c|}{4} & 0.100 & 0.577 & 4.01E-03 & 2.80E-03 & 1.06E-02 & 3.44E-03 & 3.80E-03 & \multicolumn{1}{c|}{2.19E-03} & 9.62E-04 \\
\multicolumn{1}{c|}{5} & 0.231 & -0.002 & 4.04E-04 & 5.17E-04 & 5.55E-04 & 3.61E-04 & 5.47E-04 & \multicolumn{1}{c|}{4.71E-04} & 3.70E-04 \\
\multicolumn{1}{c|}{6} & -0.137 & -0.001 & 4.16E-04 & 5.39E-04 & 5.51E-04 & 3.72E-04 & 5.41E-04 & \multicolumn{1}{c|}{4.64E-04} & 3.72E-04 \\
\multicolumn{1}{c|}{7} & -0.236 & -0.003 & 4.20E-04 & 5.48E-04 & 5.32E-04 & 3.52E-04 & 5.40E-04 & \multicolumn{1}{c|}{4.68E-04} & 3.59E-04 \\
\multicolumn{1}{c|}{8} & -0.084 & 0.206 & 1.13E-03 & 8.89E-04 & 2.76E-03 & 1.18E-03 & 1.10E-03 & \multicolumn{1}{c|}{7.67E-04} & 7.36E-04 \\
\multicolumn{1}{c|}{9} & 0.353 & -0.009 & 3.83E-04 & 5.14E-04 & 4.87E-04 & 3.18E-04 & 5.39E-04 & \multicolumn{1}{c|}{4.76E-04} & 3.33E-04 \\
\multicolumn{1}{c|}{10} & -0.093 & 0.772 & 8.39E-03 & 5.83E-03 & 2.02E-02 & 6.59E-03 & 7.95E-03 & \multicolumn{1}{c|}{4.18E-03} & 1.60E-03 \\
\multicolumn{1}{c|}{11} & -0.046 & 0.000 & 4.10E-04 & 5.29E-04 & 5.50E-04 & 3.78E-04 & 5.40E-04 & \multicolumn{1}{c|}{4.61E-04} & 3.73E-04 \\
\multicolumn{1}{c|}{12} & 0.096 & 0.830 & 1.14E-02 & 7.93E-03 & 2.58E-02 & 8.76E-03 & 1.09E-02 & \multicolumn{1}{c|}{5.78E-03} & 1.86E-03 \\
\multicolumn{1}{c|}{13} & 0.113 & 0.279 & 1.53E-03 & 1.15E-03 & 3.93E-03 & 1.52E-03 & 1.47E-03 & \multicolumn{1}{c|}{9.78E-04} & 8.58E-04 \\
\multicolumn{1}{c|}{14} & 0.132 & 0.002 & 4.12E-04 & 5.22E-04 & 5.79E-04 & 3.84E-04 & 5.48E-04 & \multicolumn{1}{c|}{4.68E-04} & 3.86E-04 \\
\multicolumn{1}{c|}{15} & -0.533 & -0.016 & 4.19E-04 & 5.95E-04 & 4.12E-04 & 2.69E-04 & 5.38E-04 & \multicolumn{1}{c|}{4.85E-04} & 2.86E-04 \\ \hline
\multicolumn{3}{c|}{Overall Error} & 1.09E-03 & 9.74E-04 & 2.25E-03 & 9.15E-04 & 1.17E-03 & \multicolumn{1}{c|}{{\ul 7.89E-04}} & \textbf{4.83E-04} \\ \hline
\end{tabular}%
}
\caption{Detailed MSE across clusters (LLaMA2-7B Layer 5 QKV-Proj Activations in Wikitext-2 inference).}
\label{tab:cluster-mse}
\end{table*}

\begin{table*}[t]
\centering
\resizebox{0.7\textwidth}{!}{%
\begin{tabular}{c|ccccccc}
\hline
\multirow{2}{*}{Layer Index} & \multicolumn{7}{c}{Data Formats} \\ \cline{2-8} 
 & NF4 & SF4 & INT4 & Asym INT4 & FP4 & Asym FP4 & Lloyd-Max \\ \hline
1  & 0.1506 & 0.1488 & 0.1259 & 0.2780 & 0.1500 & 0.0885 & 0.0701 \\
2  & 0.6235 & 0.6076 & 0.5044 & 1.2207 & 0.5081 & 0.3824 & 0.2166 \\
3  & 1.1592 & 1.0792 & 0.9745 & 2.2650 & 0.9273 & 0.7929 & 0.5121 \\
4  & 1.1049 & 1.0422 & 0.9178 & 2.1658 & 0.8747 & 0.7377 & 0.4464 \\
5  & 1.0900 & 0.9740 & 2.2500 & 0.9150 & 1.1700 & 0.7890 & 0.4830 \\
6  & 1.6423 & 1.5011 & 1.3993 & 3.1046 & 1.2913 & 1.1587 & 0.7849 \\
7  & 1.7667 & 1.6033 & 1.5124 & 3.2879 & 1.3841 & 1.2638 & 0.8648 \\
8  & 1.8086 & 1.6526 & 1.5409 & 3.3655 & 1.4081 & 1.2720 & 0.8401 \\
9  & 1.7771 & 1.6028 & 1.5328 & 3.2255 & 1.3693 & 1.2755 & 0.8917 \\
10 & 1.8307 & 1.6495 & 1.5763 & 3.2994 & 1.4067 & 1.3124 & 0.8962 \\ \hline
\end{tabular}%
}
\caption{Layer-wise quantization MSE ($\times 10^{-3}$) on LLaMA2-7B QKV projection activations.}
\label{tab:layerwise-mse}
\end{table*}

%% file: main.bbl
\begin{thebibliography}{77}
\providecommand{\natexlab}[1]{#1}

\bibitem[{AI@Meta(2024)}]{llama3modelcard}
AI@Meta. 2024.
\newblock \href {https://github.com/meta-llama/llama3/blob/main/MODEL_CARD.md} {Llama 3 model card}.

\bibitem[{AMD(2024)}]{amd_mi325}
AMD. 2024.
\newblock Amd instinct™ mi325x accelerators.
\newblock \url{https://www.amd.com/content/dam/amd/en/documents/instinct-tech-docs/product-briefs/instinct-mi325x-datasheet.pdf}.

\bibitem[{Andersch et~al.(2022)Andersch, Palmer, Krashinsky, Stam, Mehta, Brito, and Ramaswamy}]{h100}
Michael Andersch, Greg Palmer, Ronny Krashinsky, Nick Stam, Vishal Mehta, Gonzalo Brito, and Sridhar Ramaswamy. 2022.
\newblock Nvidia hopper architecture in-depth.
\newblock \url{https://developer.nvidia.com/blog/nvidia-hopper-architecture-in-depth/}.

\bibitem[{Ashkboos et~al.(2024)Ashkboos, Mohtashami, Croci, Li, Jaggi, Alistarh, Hoefler, and Hensman}]{ashkboos2024quarot}
Saleh Ashkboos, Amirkeivan Mohtashami, Maximilian~L Croci, Bo~Li, Martin Jaggi, Dan Alistarh, Torsten Hoefler, and James Hensman. 2024.
\newblock Quarot: Outlier-free 4-bit inference in rotated llms.
\newblock \emph{arXiv preprint arXiv:2404.00456}.

\bibitem[{AzureAI(2024)}]{maia}
AzureAI. 2024.
\newblock Azure maia for the era of ai: From silicon to software to systems.
\newblock \url{https://azure.microsoft.com/en-us/blog/azure-maia-for-the-era-of-ai-from-silicon-to-software-to-systems/}.

\bibitem[{Bai et~al.(2023)Bai, Bai, Chu, Cui, Dang, Deng, Fan, Ge, Han, Huang et~al.}]{bai2023qwen}
Jinze Bai, Shuai Bai, Yunfei Chu, Zeyu Cui, Kai Dang, Xiaodong Deng, Yang Fan, Wenbin Ge, Yu~Han, Fei Huang, et~al. 2023.
\newblock Qwen technical report.
\newblock \emph{arXiv preprint arXiv:2309.16609}.

\bibitem[{Bai et~al.(2024)Bai, Lv, Zhang, Lyu, Tang, Huang, Du, Liu, Zeng, Hou, Dong, Tang, and Li}]{bai-etal-2024-longbench}
Yushi Bai, Xin Lv, Jiajie Zhang, Hongchang Lyu, Jiankai Tang, Zhidian Huang, Zhengxiao Du, Xiao Liu, Aohan Zeng, Lei Hou, Yuxiao Dong, Jie Tang, and Juanzi Li. 2024.
\newblock \href {https://doi.org/10.18653/v1/2024.acl-long.172} {{L}ong{B}ench: A bilingual, multitask benchmark for long context understanding}.
\newblock In \emph{Proceedings of the 62nd Annual Meeting of the Association for Computational Linguistics (Volume 1: Long Papers)}, pages 3119--3137, Bangkok, Thailand. Association for Computational Linguistics.

\bibitem[{Bang et~al.(2024)Bang, Lee, Shim, Yang, and Chang}]{bang-etal-2024-crayon}
Jihwan Bang, Juntae Lee, Kyuhong Shim, Seunghan Yang, and Simyung Chang. 2024.
\newblock \href {https://doi.org/10.18653/v1/2024.acl-long.204} {Crayon: Customized on-device {LLM} via instant adapter blending and edge-server hybrid inference}.
\newblock In \emph{Proceedings of the 62nd Annual Meeting of the Association for Computational Linguistics (Volume 1: Long Papers)}, pages 3720--3731, Bangkok, Thailand. Association for Computational Linguistics.

\bibitem[{Bisk et~al.(2019)Bisk, Zellers, Bras, Gao, and Choi}]{bisk2019piqa}
Yonatan Bisk, Rowan Zellers, Ronan~Le Bras, Jianfeng Gao, and Yejin Choi. 2019.
\newblock \href {https://arxiv.org/abs/1911.11641} {Piqa: Reasoning about physical commonsense in natural language}.
\newblock \emph{Preprint}, arXiv:1911.11641.

\bibitem[{Bisk et~al.(2020)Bisk, Zellers, Bras, Gao, and Choi}]{Bisk2020piqa}
Yonatan Bisk, Rowan Zellers, Ronan~Le Bras, Jianfeng Gao, and Yejin Choi. 2020.
\newblock Piqa: Reasoning about physical commonsense in natural language.
\newblock In \emph{Thirty-Fourth AAAI Conference on Artificial Intelligence}.

\bibitem[{Chiang et~al.(2023)Chiang, Li, Lin, Sheng, Wu, Zhang, Zheng, Zhuang, Zhuang, Gonzalez, Stoica, and Xing}]{vicuna2023}
Wei-Lin Chiang, Zhuohan Li, Zi~Lin, Ying Sheng, Zhanghao Wu, Hao Zhang, Lianmin Zheng, Siyuan Zhuang, Yonghao Zhuang, Joseph~E. Gonzalez, Ion Stoica, and Eric~P. Xing. 2023.
\newblock \href {https://lmsys.org/blog/2023-03-30-vicuna/} {Vicuna: An open-source chatbot impressing gpt-4 with 90\%* chatgpt quality}.

\bibitem[{Chowdhery et~al.(2022)Chowdhery, Narang, Devlin, Bosma, Mishra, Roberts, Barham, Chung, Sutton, Gehrmann et~al.}]{chowdhery2022palm}
Aakanksha Chowdhery, Sharan Narang, Jacob Devlin, Maarten Bosma, Gaurav Mishra, Adam Roberts, Paul Barham, Hyung~Won Chung, Charles Sutton, Sebastian Gehrmann, et~al. 2022.
\newblock Palm: Scaling language modeling with pathways.
\newblock \emph{arXiv preprint arXiv:2204.02311}.

\bibitem[{Chung et~al.(2022)Chung, Hou, Longpre, Zoph, Tay, Fedus, Li, Wang, Dehghani, Brahma et~al.}]{chung2022scaling}
Hyung~Won Chung, Le~Hou, Shayne Longpre, Barret Zoph, Yi~Tay, William Fedus, Eric Li, Xuezhi Wang, Mostafa Dehghani, Siddhartha Brahma, et~al. 2022.
\newblock Scaling instruction-finetuned language models.
\newblock \emph{arXiv preprint arXiv:2210.11416}.

\bibitem[{Clark et~al.(2018)Clark, Cowhey, Etzioni, Khot, Sabharwal, Schoenick, and Tafjord}]{allenai:arc}
Peter Clark, Isaac Cowhey, Oren Etzioni, Tushar Khot, Ashish Sabharwal, Carissa Schoenick, and Oyvind Tafjord. 2018.
\newblock Think you have solved question answering? try arc, the ai2 reasoning challenge.
\newblock \emph{arXiv:1803.05457v1}.

\bibitem[{Dao et~al.(2022)Dao, Fu, Ermon, Rudra, and R{\'e}}]{dao2022flashattention}
Tri Dao, Daniel~Y. Fu, Stefano Ermon, Atri Rudra, and Christopher R{\'e}. 2022.
\newblock Flash{A}ttention: Fast and memory-efficient exact attention with {IO}-awareness.
\newblock In \emph{Advances in Neural Information Processing Systems (NeurIPS)}.

\bibitem[{Darvish~Rouhani et~al.(2020)Darvish~Rouhani, Lo, Zhao, Liu, Fowers, Ovtcharov, Vinogradsky, Massengill, Yang, Bittner et~al.}]{darvish2020pushing}
Bita Darvish~Rouhani, Daniel Lo, Ritchie Zhao, Ming Liu, Jeremy Fowers, Kalin Ovtcharov, Anna Vinogradsky, Sarah Massengill, Lita Yang, Ray Bittner, et~al. 2020.
\newblock Pushing the limits of narrow precision inferencing at cloud scale with microsoft floating point.
\newblock \emph{Advances in neural information processing systems}, 33:10271--10281.

\bibitem[{Darvish~Rouhani et~al.(2023)Darvish~Rouhani, Zhao, Elango, Shafipour, Hall, Mesmakhosroshahi, More, Melnick, Golub, Varatkar et~al.}]{darvish2023shared}
Bita Darvish~Rouhani, Ritchie Zhao, Venmugil Elango, Rasoul Shafipour, Mathew Hall, Maral Mesmakhosroshahi, Ankit More, Levi Melnick, Maximilian Golub, Girish Varatkar, et~al. 2023.
\newblock With shared microexponents, a little shifting goes a long way.
\newblock In \emph{Proceedings of the 50th Annual International Symposium on Computer Architecture}, pages 1--13.

\bibitem[{Dasigi et~al.(2021)Dasigi, Lo, Beltagy, Cohan, Smith, and Gardner}]{qasper}
Pradeep Dasigi, Kyle Lo, Iz~Beltagy, Arman Cohan, Noah~A. Smith, and Matt Gardner. 2021.
\newblock A dataset of information-seeking questions and answers anchored in research papers.

\bibitem[{Dettmers et~al.(2022)Dettmers, Lewis, Belkada, and Zettlemoyer}]{dettmers2022llmint8}
Tim Dettmers, Mike Lewis, Younes Belkada, and Luke Zettlemoyer. 2022.
\newblock Llm.int8(): 8-bit matrix multiplication for transformers at scale.
\newblock \emph{arXiv preprint arXiv:2208.07339}.

\bibitem[{Dettmers et~al.(2023)Dettmers, Pagnoni, Holtzman, and Zettlemoyer}]{dettmers2023qlora}
Tim Dettmers, Artidoro Pagnoni, Ari Holtzman, and Luke Zettlemoyer. 2023.
\newblock \href {https://openreview.net/forum?id=OUIFPHEgJU} {{QL}o{RA}: Efficient finetuning of quantized {LLM}s}.
\newblock In \emph{Thirty-seventh Conference on Neural Information Processing Systems}.

\bibitem[{Dosovitskiy et~al.(2021)Dosovitskiy, Beyer, Kolesnikov, Weissenborn, Zhai, Unterthiner, Dehghani, Minderer, Heigold, Gelly, Uszkoreit, and Houlsby}]{dosovitskiy2021an_vit}
Alexey Dosovitskiy, Lucas Beyer, Alexander Kolesnikov, Dirk Weissenborn, Xiaohua Zhai, Thomas Unterthiner, Mostafa Dehghani, Matthias Minderer, Georg Heigold, Sylvain Gelly, Jakob Uszkoreit, and Neil Houlsby. 2021.
\newblock \href {https://openreview.net/forum?id=YicbFdNTTy} {An image is worth 16x16 words: Transformers for image recognition at scale}.
\newblock In \emph{International Conference on Learning Representations}.

\bibitem[{Dotzel et~al.(2024)Dotzel, Chen, Kotb, Prasad, Wu, Li, Abdelfattah, and Zhang}]{dotzel2024students}
Jordan Dotzel, Yuzong Chen, Bahaa Kotb, Sushma Prasad, Gang Wu, Sheng Li, Mohamed~S. Abdelfattah, and Zhiru Zhang. 2024.
\newblock Learning from students: Applying t-distributions to explore accurate and efficient formats for llms.
\newblock \emph{International Conference on Machine Learning}.

\bibitem[{Drumond et~al.(2018)Drumond, Lin, Jaggi, and Falsafi}]{drumond2018training}
Mario Drumond, Tao Lin, Martin Jaggi, and Babak Falsafi. 2018.
\newblock Training dnns with hybrid block floating point.
\newblock \emph{Advances in Neural Information Processing Systems}, 31.

\bibitem[{Fishman et~al.(2024)Fishman, Chmiel, Banner, and Soudry}]{fishman2024scalingfp8trainingtrilliontoken}
Maxim Fishman, Brian Chmiel, Ron Banner, and Daniel Soudry. 2024.
\newblock \href {https://arxiv.org/abs/2409.12517} {Scaling fp8 training to trillion-token llms}.
\newblock \emph{Preprint}, arXiv:2409.12517.

\bibitem[{Frantar et~al.(2022)Frantar, Ashkboos, Hoefler, and Alistarh}]{frantar2022gptq}
Elias Frantar, Saleh Ashkboos, Torsten Hoefler, and Dan Alistarh. 2022.
\newblock Gptq: Accurate post-training quantization for generative pre-trained transformers.
\newblock \emph{arXiv preprint arXiv:2210.17323}.

\bibitem[{Gao et~al.(2020)Gao, Biderman, Black, Golding, Hoppe, Foster, Phang, He, Thite, Nabeshima et~al.}]{gao2020pile}
Leo Gao, Stella Biderman, Sid Black, Laurence Golding, Travis Hoppe, Charles Foster, Jason Phang, Horace He, Anish Thite, Noa Nabeshima, et~al. 2020.
\newblock The pile: An 800gb dataset of diverse text for language modeling.
\newblock \emph{arXiv preprint arXiv:2101.00027}.

\bibitem[{Gao et~al.(2021)Gao, Tow, Biderman, Black, DiPofi, Foster, Golding, Hsu, McDonell, Muennighoff, Phang, Reynolds, Tang, Thite, Wang, Wang, and Zou}]{eval-harness}
Leo Gao, Jonathan Tow, Stella Biderman, Sid Black, Anthony DiPofi, Charles Foster, Laurence Golding, Jeffrey Hsu, Kyle McDonell, Niklas Muennighoff, Jason Phang, Laria Reynolds, Eric Tang, Anish Thite, Ben Wang, Kevin Wang, and Andy Zou. 2021.
\newblock \href {https://doi.org/10.5281/zenodo.5371628} {A framework for few-shot language model evaluation}.

\bibitem[{Gliwa et~al.(2019)Gliwa, Mochol, Biesek, and Wawer}]{gliwa-etal-2019-samsum}
Bogdan Gliwa, Iwona Mochol, Maciej Biesek, and Aleksander Wawer. 2019.
\newblock \href {https://doi.org/10.18653/v1/D19-5409} {{SAMS}um corpus: A human-annotated dialogue dataset for abstractive summarization}.
\newblock In \emph{Proceedings of the 2nd Workshop on New Frontiers in Summarization}, pages 70--79, Hong Kong, China. Association for Computational Linguistics.

\bibitem[{Guo et~al.(2023)Guo, Xu, Duan, Yin, and McAuley}]{guo2023longcoderlongrangepretrainedlanguage_lcc}
Daya Guo, Canwen Xu, Nan Duan, Jian Yin, and Julian McAuley. 2023.
\newblock \href {https://arxiv.org/abs/2306.14893} {Longcoder: A long-range pre-trained language model for code completion}.
\newblock \emph{Preprint}, arXiv:2306.14893.

\bibitem[{Hendrycks et~al.(2020)Hendrycks, Burns, Basart, Zou, Mazeika, Song, and Steinhardt}]{mmlu}
Dan Hendrycks, Collin Burns, Steven Basart, Andy Zou, Mantas Mazeika, Dawn Song, and Jacob Steinhardt. 2020.
\newblock \href {https://arxiv.org/abs/2009.03300} {Measuring massive multitask language understanding}.
\newblock \emph{CoRR}, abs/2009.03300.

\bibitem[{Ho et~al.(2020)Ho, Duong~Nguyen, Sugawara, and Aizawa}]{ho-etal-2020-constructing-multihop}
Xanh Ho, Anh-Khoa Duong~Nguyen, Saku Sugawara, and Akiko Aizawa. 2020.
\newblock \href {https://doi.org/10.18653/v1/2020.coling-main.580} {Constructing a multi-hop {QA} dataset for comprehensive evaluation of reasoning steps}.
\newblock In \emph{Proceedings of the 28th International Conference on Computational Linguistics}, pages 6609--6625, Barcelona, Spain (Online). International Committee on Computational Linguistics.

\bibitem[{Horowitz(2014)}]{horowitz2014energy}
Mark Horowitz. 2014.
\newblock Energy table for 45nm process.

\bibitem[{Huang et~al.(2021)Huang, Cao, Parulian, Ji, and Wang}]{huang-etal-2021-efficient-govreport}
Luyang Huang, Shuyang Cao, Nikolaus Parulian, Heng Ji, and Lu~Wang. 2021.
\newblock \href {https://doi.org/10.18653/v1/2021.naacl-main.112} {Efficient attentions for long document summarization}.
\newblock In \emph{Proceedings of the 2021 Conference of the North American Chapter of the Association for Computational Linguistics: Human Language Technologies}, pages 1419--1436, Online. Association for Computational Linguistics.

\bibitem[{Jiang et~al.(2023)Jiang, Sablayrolles, Mensch, Bamford, Chaplot, Casas, Bressand, Lengyel, Lample, Saulnier et~al.}]{jiang2023mistral}
Albert~Q Jiang, Alexandre Sablayrolles, Arthur Mensch, Chris Bamford, Devendra~Singh Chaplot, Diego de~las Casas, Florian Bressand, Gianna Lengyel, Guillaume Lample, Lucile Saulnier, et~al. 2023.
\newblock Mistral 7b.
\newblock \emph{arXiv preprint arXiv:2310.06825}.

\bibitem[{{Joshi} et~al.(2017){Joshi}, {Choi}, {Weld}, and {Zettlemoyer}}]{2017arXivtriviaqa}
Mandar {Joshi}, Eunsol {Choi}, Daniel {Weld}, and Luke {Zettlemoyer}. 2017.
\newblock \href {https://arxiv.org/abs/1705.03551} {{triviaqa: A Large Scale Distantly Supervised Challenge Dataset for Reading Comprehension}}.
\newblock \emph{arXiv e-prints}, arXiv:1705.03551.

\bibitem[{Klimt and Yang(2004)}]{enron}
Bryan Klimt and Yiming Yang. 2004.
\newblock \href {https://api.semanticscholar.org/CorpusID:13451873} {The enron corpus: A new dataset for email classi(cid:12)cation research}.

\bibitem[{Lee et~al.(2023)Lee, Kim, Baek, Hwang, Sung, and Choi}]{enhancing}
Janghwan Lee, Minsoo Kim, Seungcheol Baek, Seok Hwang, Wonyong Sung, and Jungwook Choi. 2023.
\newblock \href {https://doi.org/10.18653/v1/2023.emnlp-main.910} {Enhancing computation efficiency in large language models through weight and activation quantization}.
\newblock In \emph{Proceedings of the 2023 Conference on Empirical Methods in Natural Language Processing}, pages 14726--14739, Singapore. Association for Computational Linguistics.

\bibitem[{Lee et~al.(2024)Lee, Park, Hong, Kim, Chang, and Choi}]{lee-etal-2024-improving-conversational}
Janghwan Lee, Seongmin Park, Sukjin Hong, Minsoo Kim, Du-Seong Chang, and Jungwook Choi. 2024.
\newblock \href {https://doi.org/10.18653/v1/2024.acl-long.612} {Improving conversational abilities of quantized large language models via direct preference alignment}.
\newblock In \emph{Proceedings of the 62nd Annual Meeting of the Association for Computational Linguistics (Volume 1: Long Papers)}, pages 11346--11364, Bangkok, Thailand. Association for Computational Linguistics.

\bibitem[{Lewis et~al.(2019)Lewis, Liu, Goyal, Ghazvininejad, Mohamed, Levy, Stoyanov, and Zettlemoyer}]{lewis2019bart}
Mike Lewis, Yinhan Liu, Naman Goyal, Marjan Ghazvininejad, Abdelrahman Mohamed, Omer Levy, Ves Stoyanov, and Luke Zettlemoyer. 2019.
\newblock Bart: Denoising sequence-to-sequence pre-training for natural language generation, translation, and comprehension.
\newblock \emph{arXiv preprint arXiv:1910.13461}.

\bibitem[{Li and Roth(2002)}]{li-roth-2002-learning-trec}
Xin Li and Dan Roth. 2002.
\newblock \href {https://aclanthology.org/C02-1150} {Learning question classifiers}.
\newblock In \emph{{COLING} 2002: The 19th International Conference on Computational Linguistics}.

\bibitem[{Lin(2004)}]{lin-2004-rouge}
Chin-Yew Lin. 2004.
\newblock \href {https://aclanthology.org/W04-1013} {{ROUGE}: A package for automatic evaluation of summaries}.
\newblock In \emph{Text Summarization Branches Out}, pages 74--81, Barcelona, Spain. Association for Computational Linguistics.

\bibitem[{Lin et~al.(2024)Lin, Xu, Wu, Cui, Zhang, Mou, Song, Sun, and Wei}]{lin2024duquantdistributingoutliersdual}
Haokun Lin, Haobo Xu, Yichen Wu, Jingzhi Cui, Yingtao Zhang, Linzhan Mou, Linqi Song, Zhenan Sun, and Ying Wei. 2024.
\newblock \href {https://arxiv.org/abs/2406.01721} {Duquant: Distributing outliers via dual transformation makes stronger quantized llms}.
\newblock \emph{Preprint}, arXiv:2406.01721.

\bibitem[{Lin et~al.(2023)Lin, Tang, Tang, Yang, Dang, and Han}]{lin2023awq}
Ji~Lin, Jiaming Tang, Haotian Tang, Shang Yang, Xingyu Dang, and Song Han. 2023.
\newblock Awq: Activation-aware weight quantization for llm compression and acceleration.
\newblock \emph{arXiv}.

\bibitem[{Liu et~al.(2023{\natexlab{a}})Liu, Li, Wu, and Lee}]{liu2023llava}
Haotian Liu, Chunyuan Li, Qingyang Wu, and Yong~Jae Lee. 2023{\natexlab{a}}.
\newblock \href {https://proceedings.neurips.cc/paper_files/paper/2023/file/6dcf277ea32ce3288914faf369fe6de0-Paper-Conference.pdf} {Visual instruction tuning}.
\newblock In \emph{Advances in Neural Information Processing Systems}, volume~36, pages 34892--34916. Curran Associates, Inc.

\bibitem[{Liu et~al.(2023{\natexlab{b}})Liu, Xu, and McAuley}]{liu2023repobenchbenchmarkingrepositorylevelcode}
Tianyang Liu, Canwen Xu, and Julian McAuley. 2023{\natexlab{b}}.
\newblock \href {https://arxiv.org/abs/2306.03091} {Repobench: Benchmarking repository-level code auto-completion systems}.
\newblock \emph{Preprint}, arXiv:2306.03091.

\bibitem[{Liu et~al.(2024{\natexlab{a}})Liu, Li, Huang, Yang, Yu, Li, Yin, lin Liu, Jin, and Bai}]{liu2024ocrbenchhiddenmysteryocr}
Yuliang Liu, Zhang Li, Mingxin Huang, Biao Yang, Wenwen Yu, Chunyuan Li, Xucheng Yin, Cheng lin Liu, Lianwen Jin, and Xiang Bai. 2024{\natexlab{a}}.
\newblock \href {https://arxiv.org/abs/2305.07895} {Ocrbench: On the hidden mystery of ocr in large multimodal models}.
\newblock \emph{Preprint}, arXiv:2305.07895.

\bibitem[{Liu et~al.(2024{\natexlab{b}})Liu, Zhao, Fedorov, Soran, Choudhary, Krishnamoorthi, Chandra, Tian, and Blankevoort}]{liu2024spinquant}
Zechun Liu, Changsheng Zhao, Igor Fedorov, Bilge Soran, Dhruv Choudhary, Raghuraman Krishnamoorthi, Vikas Chandra, Yuandong Tian, and Tijmen Blankevoort. 2024{\natexlab{b}}.
\newblock Spinquant--llm quantization with learned rotations.
\newblock \emph{arXiv preprint arXiv:2405.16406}.

\bibitem[{Lloyd(1982)}]{lloydmax}
S.~Lloyd. 1982.
\newblock \href {https://doi.org/10.1109/TIT.1982.1056489} {Least squares quantization in pcm}.
\newblock \emph{IEEE Transactions on Information Theory}, 28(2):129--137.

\bibitem[{Ma et~al.(2023)Ma, Fang, and Wang}]{ma2023llmpruner}
Xinyin Ma, Gongfan Fang, and Xinchao Wang. 2023.
\newblock Llm-pruner: On the structural pruning of large language models.
\newblock In \emph{Advances in Neural Information Processing Systems}.

\bibitem[{Marcus et~al.(1993)Marcus, Santorini, and Marcinkiewicz}]{marcus-etal-1993-building-ptb}
Mitchell~P. Marcus, Beatrice Santorini, and Mary~Ann Marcinkiewicz. 1993.
\newblock \href {https://aclanthology.org/J93-2004} {Building a large annotated corpus of {E}nglish: The {P}enn {T}reebank}.
\newblock \emph{Computational Linguistics}, 19(2):313--330.

\bibitem[{Masry et~al.(2022)Masry, Long, Tan, Joty, and Hoque}]{masry-etal-2022-chartqa}
Ahmed Masry, Do~Long, Jia~Qing Tan, Shafiq Joty, and Enamul Hoque. 2022.
\newblock \href {https://doi.org/10.18653/v1/2022.findings-acl.177} {{C}hart{QA}: A benchmark for question answering about charts with visual and logical reasoning}.
\newblock In \emph{Findings of the Association for Computational Linguistics: ACL 2022}, pages 2263--2279, Dublin, Ireland. Association for Computational Linguistics.

\bibitem[{Mathew et~al.(2021)Mathew, Karatzas, and Jawahar}]{mathew2021docvqadatasetvqadocument}
Minesh Mathew, Dimosthenis Karatzas, and C.~V. Jawahar. 2021.
\newblock \href {https://arxiv.org/abs/2007.00398} {Docvqa: A dataset for vqa on document images}.
\newblock \emph{Preprint}, arXiv:2007.00398.

\bibitem[{Merity et~al.(2016)Merity, Xiong, Bradbury, and Socher}]{merity2016pointerwiki}
Stephen Merity, Caiming Xiong, James Bradbury, and Richard Socher. 2016.
\newblock \href {https://arxiv.org/abs/1609.07843} {Pointer sentinel mixture models}.
\newblock \emph{Preprint}, arXiv:1609.07843.

\bibitem[{Nvidia(2017)}]{v100}
Nvidia. 2017.
\newblock \href {https://images.nvidia.com/content/volta-architecture/pdf/volta-architecture-whitepaper.pdf} {Nvidia tesla v100 gpu architecture}.

\bibitem[{Nvidia(2020)}]{a100}
Nvidia. 2020.
\newblock Nvidia a100 tensor core gpu architecture.
\newblock \url{https://images.nvidia.com/aem-dam/en-zz/Solutions/data-center/nvidia-ampere-architecture-whitepaper.pdf}.

\bibitem[{Nvidia(2024)}]{b100}
Nvidia. 2024.
\newblock \href {https://resources.nvidia.com/en-us-blackwell-architecture} {Nvidia blackwell architecture technical brief}.

\bibitem[{NVIDIA(2024)}]{tensorrt-llm}
NVIDIA. 2024.
\newblock Tensorrt-llm.
\newblock \url{https://github.com/NVIDIA/TensorRT-LLM}.

\bibitem[{of~the U.S. National Library~of Medicine(2023)}]{pubmeddataset}
Courtesy of~the U.S. National Library~of Medicine. 2023.
\newblock Pubmed.
\newblock \url{https://huggingface.co/datasets/ncbi/pubmed}.

\bibitem[{OpenAI(2023)}]{gpt4}
OpenAI. 2023.
\newblock Gpt-4 technical report.
\newblock \emph{arXiv preprint arXiv:2303.08774}.

\bibitem[{Rouhani et~al.(2023{\natexlab{a}})Rouhani, Garegrat, Savell, More, Han, Zhao, Hall, Klar, Chung, Yu, Schulte, Wittig, Bratt, Stephens, Milanovic, Brothers, Dubey, Cornea, Heinecke, Rodriguez, Langhammer, Deng, Naumov, Micikevicius, Siu, and Verrilli}]{ocp}
Bita~Darvish Rouhani, Nitin Garegrat, Tom Savell, Ankit More, Kyung-Nam Han, Ritchie Zhao, Mathew Hall, Jasmine Klar, Eric Chung, Yuan Yu, Michael Schulte, Ralph Wittig, Ian Bratt, Nigel Stephens, Jelena Milanovic, John Brothers, Pradeep Dubey, Marius Cornea, Alexander Heinecke, Andres Rodriguez, Martin Langhammer, Summer Deng, Maxim Naumov, Paulius Micikevicius, Michael Siu, and Colin Verrilli. 2023{\natexlab{a}}.
\newblock \href {https://www.opencompute.org/documents/ocp-microscaling-formats-mx-v1-0-spec-final-pdf} {Ocp microscaling formats (mx) specification}.

\bibitem[{Rouhani et~al.(2023{\natexlab{b}})Rouhani, Zhao, More, Hall, Khodamoradi, Deng, Choudhary, Cornea, Dellinger, Denolf et~al.}]{rouhani2023microscaling}
Bita~Darvish Rouhani, Ritchie Zhao, Ankit More, Mathew Hall, Alireza Khodamoradi, Summer Deng, Dhruv Choudhary, Marius Cornea, Eric Dellinger, Kristof Denolf, et~al. 2023{\natexlab{b}}.
\newblock Microscaling data formats for deep learning.
\newblock \emph{arXiv preprint arXiv:2310.10537}.

\bibitem[{Sakaguchi et~al.(2019)Sakaguchi, Bras, Bhagavatula, and Choi}]{sakaguchi2019winogrande}
Keisuke Sakaguchi, Ronan~Le Bras, Chandra Bhagavatula, and Yejin Choi. 2019.
\newblock \href {https://arxiv.org/abs/1907.10641} {Winogrande: An adversarial winograd schema challenge at scale}.
\newblock \emph{Preprint}, arXiv:1907.10641.

\bibitem[{See et~al.(2017)See, Liu, and Manning}]{see-etal-2017-get-cnn}
Abigail See, Peter~J. Liu, and Christopher~D. Manning. 2017.
\newblock \href {https://doi.org/10.18653/v1/P17-1099} {Get to the point: Summarization with pointer-generator networks}.
\newblock In \emph{Proceedings of the 55th Annual Meeting of the Association for Computational Linguistics (Volume 1: Long Papers)}, pages 1073--1083, Vancouver, Canada. Association for Computational Linguistics.

\bibitem[{Shao et~al.(2024)Shao, Chen, Zhang, Xu, Zhao, Li, Zhang, Gao, Qiao, and Luo}]{shao2024omniquant}
Wenqi Shao, Mengzhao Chen, Zhaoyang Zhang, Peng Xu, Lirui Zhao, Zhiqian Li, Kaipeng Zhang, Peng Gao, Yu~Qiao, and Ping Luo. 2024.
\newblock \href {https://openreview.net/forum?id=8Wuvhh0LYW} {Omniquant: Omnidirectionally calibrated quantization for large language models}.
\newblock In \emph{The Twelfth International Conference on Learning Representations}.

\bibitem[{Singh et~al.(2019)Singh, Natarjan, Shah, Jiang, Chen, Batra, Parikh, and Rohrbach}]{singh2019towardstextvqa}
Amanpreet Singh, Vivek Natarjan, Meet Shah, Yu~Jiang, Xinlei Chen, Dhruv Batra, Devi Parikh, and Marcus Rohrbach. 2019.
\newblock Towards vqa models that can read.
\newblock In \emph{Proceedings of the IEEE Conference on Computer Vision and Pattern Recognition}, pages 8317--8326.

\bibitem[{Talmor et~al.(2019)Talmor, Herzig, Lourie, and Berant}]{talmor2019commonsenseqa}
Alon Talmor, Jonathan Herzig, Nicholas Lourie, and Jonathan Berant. 2019.
\newblock \href {https://arxiv.org/abs/1811.00937} {Commonsenseqa: A question answering challenge targeting commonsense knowledge}.
\newblock \emph{Preprint}, arXiv:1811.00937.

\bibitem[{Touvron et~al.(2023)}]{touvron2023llama2}
Hugo Touvron et~al. 2023.
\newblock \href {https://arxiv.org/abs/2307.09288} {Llama 2: Open foundation and fine-tuned chat models}.
\newblock \emph{Preprint}, arXiv:2307.09288.

\bibitem[{Xiao et~al.(2022)Xiao, Lin, Seznec, Demouth, and Han}]{xiao2022smoothquant}
Guangxuan Xiao, Ji~Lin, Mickael Seznec, Julien Demouth, and Song Han. 2022.
\newblock Smoothquant: Accurate and efficient post-training quantization for large language models.
\newblock \emph{arXiv preprint arXiv:2211.10438}.

\bibitem[{Yang et~al.(2024)Yang, Kim, and Kim}]{yang2024mitigatingquantizationerrorsactivation}
Jaewoo Yang, Hayun Kim, and Younghoon Kim. 2024.
\newblock \href {https://arxiv.org/abs/2405.14428} {Mitigating quantization errors due to activation spikes in glu-based llms}.
\newblock \emph{Preprint}, arXiv:2405.14428.

\bibitem[{Yang et~al.(2018)Yang, Qi, Zhang, Bengio, Cohen, Salakhutdinov, and Manning}]{yang2018hotpotqadatasetdiverseexplainable}
Zhilin Yang, Peng Qi, Saizheng Zhang, Yoshua Bengio, William~W. Cohen, Ruslan Salakhutdinov, and Christopher~D. Manning. 2018.
\newblock \href {https://arxiv.org/abs/1809.09600} {Hotpotqa: A dataset for diverse, explainable multi-hop question answering}.
\newblock \emph{Preprint}, arXiv:1809.09600.

\bibitem[{Zhang et~al.(2025{\natexlab{a}})Zhang, Huang, Zhang, Wei, Zhu, and Chen}]{zhang2024sageattention2}
Jintao Zhang, Haofeng Huang, Pengle Zhang, Jia Wei, Jun Zhu, and Jianfei Chen. 2025{\natexlab{a}}.
\newblock Sageattention2: Efficient attention with thorough outlier smoothing and per-thread int4 quantization.
\newblock In \emph{International Conference on Machine Learning (ICML)}.

\bibitem[{Zhang et~al.(2025{\natexlab{b}})Zhang, Wei, Zhang, Xu, Huang, Wang, Jiang, Zhu, and Chen}]{zhang2025sageattention3}
Jintao Zhang, Jia Wei, Pengle Zhang, Xiaoming Xu, Haofeng Huang, Haoxu Wang, Kai Jiang, Jun Zhu, and Jianfei Chen. 2025{\natexlab{b}}.
\newblock Sageattention3: Microscaling fp4 attention for inference and an exploration of 8-bit training.
\newblock \emph{arXiv preprint arXiv:2505.11594}.

\bibitem[{Zhang et~al.(2025{\natexlab{c}})Zhang, Wei, Zhang, Zhu, and Chen}]{zhang2025sageattention}
Jintao Zhang, Jia Wei, Pengle Zhang, Jun Zhu, and Jianfei Chen. 2025{\natexlab{c}}.
\newblock Sageattention: Accurate 8-bit attention for plug-and-play inference acceleration.
\newblock In \emph{International Conference on Learning Representations (ICLR)}.

\bibitem[{Zhang et~al.(2024{\natexlab{a}})Zhang, Li, Zhang, Pu, Cahyono, Hu, Liu, Zhang, Yang, Li, and Liu}]{zhang2024lmmsevalrealitycheckevaluation}
Kaichen Zhang, Bo~Li, Peiyuan Zhang, Fanyi Pu, Joshua~Adrian Cahyono, Kairui Hu, Shuai Liu, Yuanhan Zhang, Jingkang Yang, Chunyuan Li, and Ziwei Liu. 2024{\natexlab{a}}.
\newblock \href {https://arxiv.org/abs/2407.12772} {Lmms-eval: Reality check on the evaluation of large multimodal models}.
\newblock \emph{Preprint}, arXiv:2407.12772.

\bibitem[{Zhang et~al.(2022)Zhang, Roller, Goyal, Artetxe, Chen, Chen, Dewan, Diab, Li, Lin, Mihaylov, Ott, Shleifer, Shuster, Simig, Koura, Sridhar, Wang, and Zettlemoyer}]{opt}
Susan Zhang, Stephen Roller, Naman Goyal, Mikel Artetxe, Moya Chen, Shuohui Chen, Christopher Dewan, Mona Diab, Xian Li, Xi~Victoria Lin, Todor Mihaylov, Myle Ott, Sam Shleifer, Kurt Shuster, Daniel Simig, Punit~Singh Koura, Anjali Sridhar, Tianlu Wang, and Luke Zettlemoyer. 2022.
\newblock \href {https://doi.org/10.48550/ARXIV.2205.01068} {Opt: Open pre-trained transformer language models}.
\newblock \emph{arXiv preprint}.

\bibitem[{Zhang et~al.(2024{\natexlab{b}})Zhang, Zhang, Cao, Du, Wei, Cao, and Xu}]{zhang2023afpq}
Yijia Zhang, Sicheng Zhang, Shijie Cao, DaYou Du, Jianyu Wei, Ting Cao, and Ningyi Xu. 2024{\natexlab{b}}.
\newblock \href {https://doi.org/10.18653/v1/2024.findings-acl.3} {{AFPQ}: Asymmetric floating point quantization for {LLM}s}.
\newblock In \emph{Findings of the Association for Computational Linguistics ACL 2024}, pages 28--36, Bangkok, Thailand and virtual meeting. Association for Computational Linguistics.

\bibitem[{Zheng et~al.(2023)Zheng, Chiang, Sheng, Zhuang, Wu, Zhuang, Lin, Li, Li, Xing, Zhang, Gonzalez, and Stoica}]{zheng2023judging}
Lianmin Zheng, Wei-Lin Chiang, Ying Sheng, Siyuan Zhuang, Zhanghao Wu, Yonghao Zhuang, Zi~Lin, Zhuohan Li, Dacheng Li, Eric Xing, Hao Zhang, Joseph~E. Gonzalez, and Ion Stoica. 2023.
\newblock \href {https://openreview.net/forum?id=uccHPGDlao} {Judging {LLM}-as-a-judge with {MT}-bench and chatbot arena}.
\newblock In \emph{Thirty-seventh Conference on Neural Information Processing Systems Datasets and Benchmarks Track}.

\end{thebibliography}
